# Translation, Sentiment and Voices: A Computational Model to Translate and Analyze Voices from Real-Time Video Calling

## Aneek Barman Roy

School of Computer Science and Statistics

Department of Computer Science

Thesis submitted for the degree of Master in Science

Trinity College Dublin

2019

# List of Tables

# List of Figures

# List of Acronyms

API Application Programming Interface

ASR Automatic Speech Recognition

BLEU BiLingual Evaluation Understudy

CNN Convolutional Neural Network

EBMT Example Based Machine Translation

GNMT Google Neural Machine Translation

GRU Grated recurring Units

LSTM Long Short-Term Memory

NLP Natural Language Processing

NLTK Natural Language Tool Kit

NMT Neural Machine Translation

OpenNMT Open Neural Machine Translation

RBMT Rule Based Machine Translation

RNN Recurring Neural Network

SMT Statistical Machine Translation

# Contents

# 1

# Introduction

This chapter starts briefly mentions about the background and motivation which were needed for this work. In Section 1.2 and 1.3, the research question and objectives are set. Section 1.4 discusses the research challenges that could be involved while a technical overview is provided in Section 1.5. Finally, the chapter concludes in Section 1.6 where the structure of this thesis has been explained.

## 1.1 Motivation and Background

The current decade has witnessed the rise of multi-media chat applications and smart-phone users. Initially the boom of chat applications happened in the late 2010s. With internet quickly becoming an easy access to many, voice calling over internet has been slowly gaining momentum. Individuals can connect to their loved ones very quickly with a touch of their smartphone screen.

Even though the internet helped to connect people who are far away from each other, individuals still struggle during a communication. Many individuals have been engaging in video communication across the world in different languages. Since there are many languages spoken around the world, it becomes very hard for individuals from a developing country to converse if English is not his or her first language. For example, an individual whose first language is Bengali might struggle to speak in French with another individual. In Europe, such cases are still prevalent. Often individuals struggle to communicate with each other if both of them belong to different linguistics groups i.e. French and English. Interestingly, French and English have same origin i.e. Latin.

The decade saw also the emergence of language translation using neural networks as well. Till the mid-2020s, machine language relying on statistics were in use. The introduction of a neural approach changed the research scenario in natural language processing (NLP) and machine translation (MT).

With more data being generated in audio and visual forms, it has become a need and a challenge to analyse such information for many researchers from academia and industry. The availability of video chat corpora is limited as organizations protect user privacy and ensure data security. It was due to this reason, a video chat application was conceptualized which can give the speaker a choice to record his or her session.

The understanding of human emotions from video chat sessions formed a strong background in this work. With the help of the extracted speeches, it is possible to determine the behavioural characteristics of a person. The polarity and loudness were chosen to be the parameters for such a determination. This is due to the common understanding that vocal loudness influences the behaviour of a person. It will be interesting to find out if polarity i.e. the sentiment and vocal intensity are corelated to each other. Meanwhile, with the introduction of Grated Recurring Units (GRU) in a LSTM -era i.e. Long Short-Term Memory, machine learning gained more momentum as GRU was dynamic in nature. The language translation model of this work is influenced by the high-speed workings of GRU. The translation model takes a recurrent neural approach and converts English sentences to French. Since the source data point is textual i.e. it is dependent on other textual data points, the RNN model has been chosen.

On the whole, this work aims to understand human nature in a video session with the help of voice and sentiment analysis. The goal is to ultimately empower the audio-visual communication systems to translate spoken languages based on previous emotional learnings. Since the larger picture is out of scope for this work, five objectives are drawn for the project and mentioned in the next section. Additionally, a research question has been set as well.

## 1.2    Research Question

The thesis centres around the question how the creation of a video chat application and a language translation model can translate the voice messages of actors and understand human behaviour

## 1.3    Research Objectives

In order to address the research question, we have established five research objectives:

1. To create an audio-visual communication system which has the ability to record the speech of a speaker

2. To create a small speech corpus from the audio-visual communication system

3. To understand the behaviour of speakers in a video session

4. To create a neural translation system to convert English source languages to French

5. To evaluate and run the translation model with the speech corpus

## 1.4    Research Challenges

In addition to the research objectives, there are specific challenges which are possible in the research areas of video processing, speech analysis and natural language processing. The research challenges can be addressed as:

1. Natural language Processing is a vast area of research where extensive study is required. English and French were chosen as source and target languages as both have same origin i.e. Latin. The concept of machine translation using neural network is close to half-a-decade old. Although numerous works have been done language translations, the challenge is to test the neural network model on the real-time audio-speeches.

2. The corpus for a real-time video chat speech database is limited. It is due to reason of maintaining user privacy and data security. The challenge is to create a database of audio-speeches which can be analysed for this work.

3. Many researchers have been using Long Short-Term Memory(LSTM) for their neural layers. Here the challenge is to use a Grated Recurrent Unit (GRU) in neural network layers which has one logic gate less than the former.

4. In order to obtain the speeches of speakers from the audio-visual communication systems, there is a challenge.

5. There is a missing link in the area of sentiment analysis and analysis of real-time audio-visuals. Thus, the challenge is to bridge the gap between the two research areas

## 1.5 Technical Overview

The focus of this work is on the creation of an audio-visual communication system and the computational model to translate and analyse voices from the system. There are three phases involved in this work, namely:

- an audio-visual communication system
- speech analysis of voices to understand human behaviour
- neural translation system to convert English sentences to French

The availability of video chat corpora is limited as organizations protect user privacy and ensure data security. For this reason, an audio-visual communication system (VidALL) has been developed where a speaker can record his or her chat session. The audio-speeches were extracted and converted to textual form. Thus, the audio samples were collected which led to the creation of a small corpus consisting of audio-clip files.

To understand human nature while answering a video call, an analysis was conducted on the voices where polarity and vocal intensity were considered as parameters. Since polarity signify sentiment scores, the audio files were converted to textual form.

Simultaneously, a translation model using a neural approach was developed to translate English sentences to French. Since the work on translation model is based on textual data, the use of a Recurrent Neural Network (RNN) is recommended.

While the translation model is check on its accuracy, the sentence correctness can be determined with BLEU score and a sentence comparator.

## 1.6    Structure of Thesis

The structure of this thesis has been discussed below:

- Chapter 1 is an Introduction to this dissertation which provides an overview of the project. It discusses the background work and motivation which resulted in the cultivation of this work. It provides a research question along with its objectives and challenges. The chapter also discusses the key conclusions from this dissertation. (Introduction)

- Chapter 2 studies the Start of Art systems that exist in video chat application and language translation. It presents a summary of the works that have been conducted in the past in language translation, sentiment analysis, and sound analysis. Additionally, it determines the approaches that are required for this work. Finally, the chapter compares the objectives in this work with the literature works of several authors.  (State of Art)

- Chapter 3 presents a description of the different phases that are present in this work. It proposes a methodology which was eventually followed to conduct this research work. Additionally, it talks about the design approaches which were considered to meet the research objective. The failure of some design approaches has been reasoned and the roadblocks to the final design have been mentioned. (Design and Methodology)

- Chapter 4 gives a detailed discussion on the Implementation strategies which were undertaken for this piece of work. The chapter discusses the technologies that were used and interlinks the different phases of this dissertation. (Implementation)

- Chapter 5 presents the observations that have been gathered in this research. The chapter discusses the results and evaluate the research work on the basis of research objectives, chosen models and research area. It gives reasoning of the metrics that have been chosen for different modelling purposes. (Evaluation)

- Chapter 6 highlights the contributions that have been made in this research work. The limitations and roadblocks this work incurred have also been discussed in this chapter. Additionally, it discusses the possible approaches that can be taken in the future to improve this work. Finally, the chapter ends with a summary of this dissertation along with some comments. (Conclusion)

# 2

# State of Art

The State of Art in audio-visual communication systems, detection of sentiment in speeches, and language translation have been discussed in this chapter. The chapter discusses the existing methodologies and research areas of this work and concludes with a literature review.

## 2.1 Introduction

The current decade has witnessed the rise of smart-phone users and chat applications. With internet becoming an easy access to many users, voice calling over internet is slowly gaining momentum. Additionally, there has been immense development in video chat applications in this decade. Individuals can engage in video communication across the world. Interestingly, this decade saw the emergence of language translation using neural networks. With more data being generated in audio and visual forms, it has become a need and a challenge to analyse such information for many researchers from academia and industry.

This thesis involves analysis and translation of speeches of video chat applications. Firstly, a video chat application was developed from which speeches were extracted. This was followed by an analysis of voices of the speakers based on two aspects – loudness and sentiment. Simultaneously, a translation model using a neural approach was developed to translate English sentences to French.

This chapter starts with a discussion on State of Art systems and eminent works in language translation, audio-visual communication systems, and sentiment analysis. A brief explanation on the evolution of machine language translation has been given in Section 2.2. The sub-section also includes a discussion on the existing neural machine translation systems. Section 2.3 outlines the different APIs and technologies which help to make an audio-visual communication system.

In Section 2.4, important works of sentiment analysis have been discussed. Since our work involved a lot of methodologies and a deep understanding of the research areas, several works have been studied. For this purpose, a comparative analysis of works in sentiment and emotion analysis has been mentioned in Table 2.1 whereas Table 2.2 provides a tabular description of works in the field of machine language translation. Based on our research areas and literary works studied, a literature review is provided in Section 2.5. Finally, the chapter concludes with a section where we compare the existing works to our project requirements.

## 2.2    Language Translation

Language translation in machines may be defined as the process of using computing power to translate a source language to a target language. In 1949, Warren Weaver first proposed the idea of translating languages using machines [58]. The progress in the modern-day Natural Language Processing (NLP) is a result of several theories, researches, and executions.

### 2.2.1    Background

The 1970s saw the emergence of Rule Based Machine Translation (RBMT) where the systems comprised of bilingual dictionary and a set of semantic rules defined for each language. Such systems were followed by the introduction of Example Based Machine Translation (EBMT) where ready-made phrases were used instead of recurrent translation. EBMT focused on the importance of feeding translations to the system and tried to eradicate the long process of rule formations.

The true revolution in translation research came with the arrival of Statistical Machine Translation (SMT) in early 1990s. SMT ran analysis of similar texts in bilingual and learnt linguistics patterns guided by statistics. The SMT era comprised of word-based SMT(WBSMT), phrase-based SMT (PBSMT) and syntax-based SMT (SBSMT) in a chronological order.

The WBSMT followed a classical approach of dividing sentences into words and counted the statistics. It followed "the bag of words" approach where a word was converted into multiple words. A limitation of this approach was that the converse did not turn out to be true. Since the approach did not take into account the order of words, another model was conceptualised. The new approach memorized the order in which certain words usually took place at the output and rearranged the words at the transitional step. To improve the model, the concept of null token was introduced where the machine considered the importance of a new term. Additionally, the upgraded model picked the correct particle or term for every token-word arrangement. However, it was observed that adjectives and nouns switched cases very often. Thus, there was a need for reordering which paved the way for a new model in WBSMT based on word alignment.

However, WBSMT was not trained to identify several cases including homonym and gender.

The limitations and complexity of several models of WBSMT were solved to some extent with the introduction of n-grams concept. n-grams denoted a sequence of n terms in a row thereby paving way to the formation of PBSMT. For instance, if "as" is considered as unigram for the term "as", "as soon" and "as soon as" can be considered as bigram and trigram respectively, and so on. The method took care of the basic translation principles which were defined in the word-based method i.e. reordering, statistics and vocabular tricks. Evidently, this approach improved accuracy in translations. However, if human evaluation took place in terms of structure of the sentence or linguistics, the quality of the output sentence reduced significantly.

*"Every time I fire a linguist, the performance of the speech recognizer goes up."- joked Frederick Jelinek[1]. (Hirschberg et al. [16])*

SBSMT is based on the learning of different parts of speech and building a sentence tree on them. According to Yamada et al. [62], the machine uses the tree to learn and transform the syntactic units present between different languages. The rest of the units of the speech is translated by phrases or words.

The recent emergence of Neural Machine Translation (NMT) is shifting paradigms and has generated high interests from academia and industry. The NMT is an end-to-end network which involves building of neural models to translate messages from one language to another. For instance, if a German sentence is given as input, the NMT will train in a way to maximise the likelihood of a target sentence which is in English. The neural approach does not depend on any linguistic information externally. Kalchbrenner and Blunsom [20] and Sutskever et al. [49] came up with their versions of an encoder-decoder model using neural approaches. While Kalchbrenner and Blunsom used Convolutional Neural Network (CNN) and Recurrent Neural Network (RNN), Sutskever came up with the concept of Long-Short Term Memory (LSTM). The novel approach gathered momentum with the framework of Bahdanau et al. [3].

A part of this work is based on a neural translation model which follows a supervised learning approach where the model is trained on English and French sentence. This section will discuss the various state of art works which has happened in the NMT.

---

[1] https://en.wikipedia.org/wiki/Frederick_Jelinek

### 2.2.2    Google Neural Machine Translation (GNMT)

In 2016, Google (Wu et al. [61]) came up with sequence-to-sequence neural translation model which followed the framework suggested by Sutskever (2016) combined with an attention network (Bahdanau et al., [3]). In addition to the attention network, the model consisted of two more networks – encoder network and decoder network. The encoder module connects the attention network with the decoder network so that the latter can focus on different areas of the input or source sentence during the decoding phase. The GNMT architecture stores each symbol of a sentence into one vector unit and decodes to one symbol at a time until the system reaches the end-of-sentence (EOS) symbol.

The GNMT consists of a deep Long-Short Term Memory (LSTM) network with eight layers of encoders and decoders. There are two RNNs present in the GNTM architecture where one RNN accepts the source sentence and the second RNN generates the translated target sentence. In Figure 2.1, the architecture of Google Neural Translation Machine is shown.

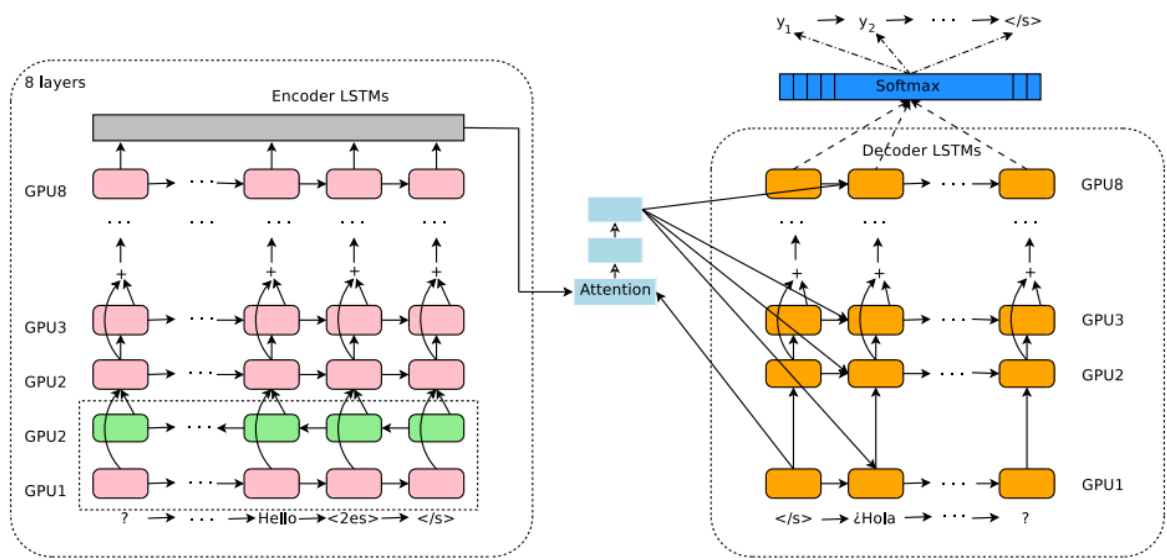

Figure 2.1: Google Neural Machine Translation Architecture (Wu et al.,2016)

From Figure 2.1, we can see that there are eight layers of Graphics Processing Units (GPU) where encoders are present in each layer. In order to fasten the training process of the model, GPUs are used. In the encoder LSTMs, there is a presence of one bi-directional layers in the bottom while the remaining seven layers are uni-directional in nature. The source sentences pass through the encoder layers to the attention network and are distributed to the decoder LSTMs. It can be observed that only GPU1(from Figure 2.1) obtain the repetitive attention context which is then sent to the other decoder layers. The researchers have partitioned the activation function softmax for the output layer and placed on different GPUs.

In order to train their model, the researchers at Google used two large corpora from WMT 2014[2] which are available in public. The WMT $En \rightarrow Fr$ was used to translate English sentences to French while the WMT $En \rightarrow De$ corpus was used to translate English sentences to German. Additionally, the authors have used Google's translation corpora. While the former corpus contains 36 million sentence pairs, the latter has 5 million sentence pairs on the whole.

The GNMT model was trained in TensorFlow[3]. Adam's Optimizer Algorithm (Kingma et al., [25]) as the learning algorithm which followed as learning rate of 0.0002. In addition to this, TensorFlow's SGD learning algorithm was used which followed a rate of 0.5. To conduct the experiments, several word-piece models viz. word-only, character-based and mixed-character, were used.

The researchers evaluated the correctness of the translated sentences using BLEU scores and human evaluation. The BLEU metric scores ranged from 0 to 6 where 6 represented a perfectly translated sentence. The side-by-side evaluations from the data has been visualized in Figure 2.2.

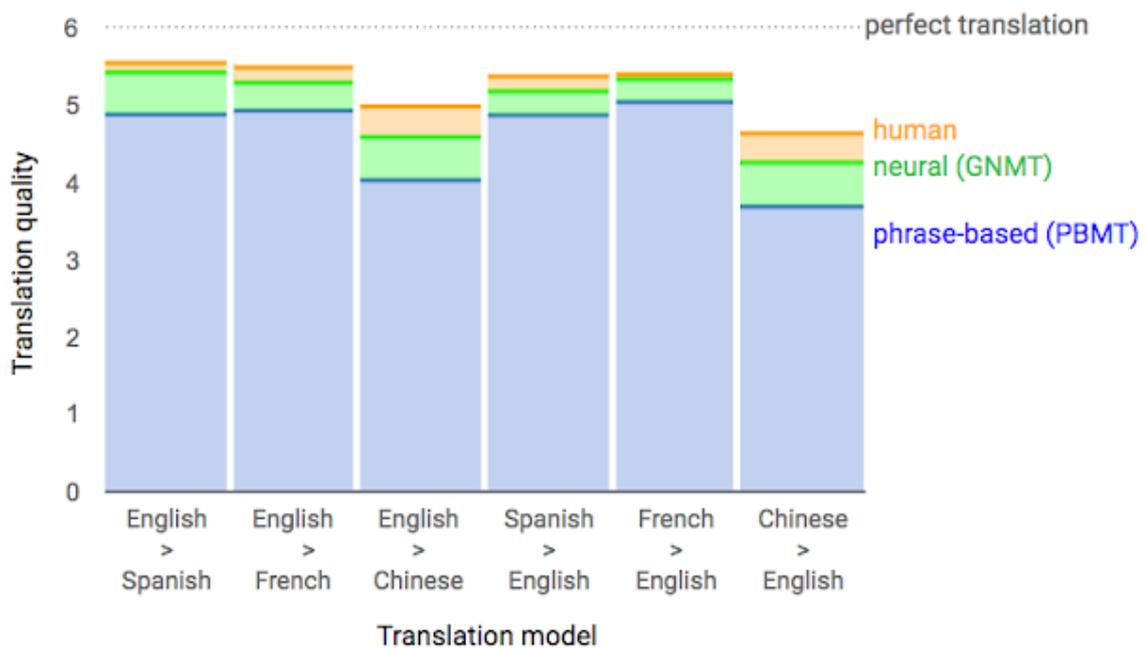

Figure 2.2: Side-by-side Evaluation of Google Neural Translation Model (Wu et al.,2016)

The GNMT solved some limitations which were present in NMT, i.e.:

- Lack of robustness which were prevalent in previous systems

- Solved the occurrence of rare words in source sentences

### 2.2.3   Open Neural Machine Translation (OpenNMT)

In 2017, SYSTRAN[4] and Harvard University[5] came together to jointly develop OpenNMT (Klein et al., [27]) which is open source in nature. The research aim of this framework was to ensure modularity and readability of the model, give preference to training-test efficiency and to provide support to significant researches. In contrast and comparison to existing translation systems where licenses are restricted, the researchers gave support to the production users with OpenNMT. The system is a successor of *seq2seq-attn*[6] translation system.

The source words entering into the OpenNMT are stored as one vector unit for each word which are then fed into the RNN. In Figure 2.3, the diagram of the OpenNMT architecture is provided.

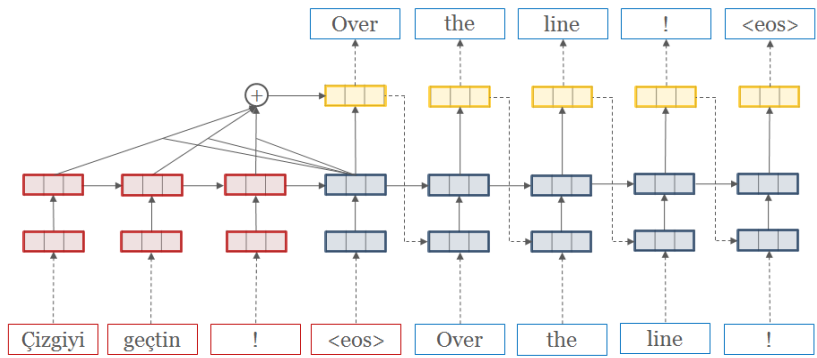

Figure 2.3: Open Neural Translation Machine Architecture (Klein et al., 2017)

The source words are denoted by red colour which enters the RNN initially. The target RNN which is blue in colour is initialized by the final time step when end-of-sentence (EOS) symbol is reached. The attention network (*denoted by +*) is implemented on the source RNN at each time step of the target. It is then combined with the hidden state to generate a prediction of the following word. The prediction is then sent to the target RNN.

In order to train their model, the researchers used the WMT 2015[7] corpora which contains 15.8 million sentence pairs of English and German. BLEU metric was used to evaluate OpenNMT

---

translations which scored 17.60 and 19.34 when the model ran on test data of vocabulary sizes 50K and 32K respectively. In both the cases, OpenNMT secured better scores when compared to Nematus[8].

### 2.2.4    Nematus

In 2017, Sennrich et al. [51] came up with Nematus, an open source NLP toolkit where the encoder-decoder architecture was similar to the one suggested by Bahdanau et al. [3]. The toolkit was based on Theano (Team [54]) and has been implemented in Python. In comparison to Bahdanau's work, Sennrich generalized the embedding layer of encoder to include random features besides the word feature.

Nematus produced some implementational changes in its neural model although any extra biases were removed from its embedding layers. The hidden state of decoder was initialized with the average of source annotation instead of annotating at the final position of the encoder. The encoder gave importance to the parts-of-speech (POS) tags, morphological features and linguistic dependency labels in the source sentences which were taken as input. It is to be noted that this model implemented a conditional Gated Recurrent Unit (GRU) with the conditional layer.

In order to train the framework, the researchers used the *En →De* and *En→Ro* corpora from WMT 2016[9]. While the former corpus contains 4.2 million sentence pairs, the latter has 2.2 million sentence pairs on the whole. The researchers ran BLEU and CHRF3[10] tests on the learning model and observed promising results for both the corpus. It was concluded that linguistics features are positively correlated with perplexity on the training data.

## 2.3    Audio-Visual Communication Systems

Video calling has transformed the mode of communication between two individuals around the world. The high-tech companies like Google, Microsoft, Facebook are using video-calling features through their products viz. Google Duo, Skype, WhatsApp respectively. An increase in bandwidth and a greater number of video chat applications gave rise to more people adopting the video chat culture. With the help of video chat APIs (Application Protocol Interface) in terms of providing technical infrastructure, it has become easier to design a basic audio-visual communication system. In such cases, the communication between the end users take place over a Voice Over Internet Protocol (VOIP). Since, our work involves analysis of audio-visual sessions from video chat applications, several APIs have been studied to make the application.

---

[8] https://github.com/EdinburghNLP/nematus
[9] http://www.statmt.org/wmt16/
[10] https://aclweb.org/anthology/W15-3049

This section discusses some Start of Art video chat APIs, namely PubNub, Sinch, and TokBox, whose technical infrastructure services can be used for our work.

### 2.3.1    PubNub

PubNub[11] is a signalling protocol service which is reliable in nature and provides a low latency messaging. This signalling helps to establish a peer-to-peer connection. PubNub co-ordinates the communication between two users and transfer the metadata before any call. It is to be noted that a video chat application can be developed if PubNub partners with WebRTC. The latter i.e. WebRTC makes the communication of video, audio and data possible between different browsers.

Although PubNub and WebRTC mostly worked for video chat applications in browsers, a cross-platform video chat is now possible due to the recent release of a WebRTC software developing kit (SDK) which provides support to android devices. Thus, PubNub does not provide a fully hosted WebRTC service. Additionally, it is not possible to store any audio or video stream if PubNub services are used. It can be seen that the limitations of PubNub does not make it suitable for developing a video chat application. At the same time, the infrastructure of PubNub is highly reliable and ensures an enterprise grade security.

### 2.3.2    Sinch

Sinch provides a platform where a developer's application can be enriched with voice, video and real-time communication. Additionally, it provides verification services and optimizes the audio quality along with video selection. The adaptive nature of a codec selection and a WebRTC API ensures an optimum audio quality. In total, Sinch supports VOIP, Session Initiation Protocol (SIP) and Public Switched Telephone Network (PSTN) across many platforms viz. android, iOS, Web Applications. The various highlighted features of Sinch are as follows:

- Signalling Firewall
- SMS Messaging
- RCS Messaging
- Verification
- Voice Calling
- Personalized Messaging

---

[11] https://www.pubnub.com

Besides app-to-app calling, Sinch offer services to support app-to-phone calling and phone-to-phone calling. However, the developers have to pay for the services although there are no hidden charges.

### 2.3.3  TokBox

TokBox ensures an easy embedding of video and voice into web and mobile applications. The OpenTok library of TokBox is responsible for providing core functionalities like:

- Connection to a session
- Publishing the stream to the session
- Subscribing to streams in the session

TokBox complements WebRTC like other APIs and supports android, iOS, web applications etc. The OpenTok video platform of TokBox helps to supply enterprise grade WebRTC. Since it is based on Java, many developers find this platform easy to use and integrate with their android applications. In addition to these, TokBox provides support for server-side SDKs (Software Development Kit) and REST API. As per the industry standards, WebRTC is in demand among all other technologies to create video chat applications. In terms of costs, TokBox is cheaper compared to the above-mentioned APIs. In addition to this, TokBox offers free credits to a developer initially.

Due to the limitations of PubNub, Sinch and TokBox can be considered to develop audio-visual applications. However, the services offered by Sinch is more in number for voice support than audio-visual support. On the basis of support and security provided by TokBox, the latter is a better choice for our research aims. Additionally, the developers prefer to use TokBox over Sinch.

## 2.4  Sentiments in Audio-Visuals

Sentiment analysis can a vital role in understanding human behaviour by analysing their conversations. It measures the opinion of individuals through computational linguistics, natural language processing, and text analysis. For this thesis in this context, the idea is to analyse the speeches generated from voice chat application and understand the behaviour of the speakers based on their loudness and polarity. Polarity is a measure to understand the sentiments. The polarity of voice can be derived if the speeches are converted to text. In this thesis, we converted the speeches to text using Google's Cloud Speech-to-Text API. Based on the transcriptions of speeches, the sentiment score is obtained. To measure the sentiment score, a study of different Natural Language Processing (NLP) frameworks have been studied.

This section discusses the different NLP frameworks that can be used to determine the sentiment of a speech transcription.

### 2.4.1 Natural Language Tool Kit (NLTK)

Natural Language Tool Kit or NLTK (Loper et al., [33], Bird et al., [5]) is a popular python-based NLP framework for language and text processing. It has interfaces to more than fifty corpora and lexical resources and can be used for several tasks like stemming, tokenization, POS tagging, lemmatization, etc. Additionally, the framework supports a wide range of third-party extensions as well as language supports for many listed libraries. One such lexicon of NLTK that is dedicated to sentiments conveyed in social media as well is Valence Aware Dictionary and sEntiment Reasoner (Hutto et al.,[18]) or VADER. Besides giving a polarity score (positive or negative), VADER also demonstrates the percentage of negative or positive of a sentiment.

The algorithm used in this lexicon gives sentiment scores in a class of four: negative, positive, neutral and compound. The proportion of sentiment scores of a text is given in the negative, positive, and neutral classes. For instance, for a source sentence: "*The phone is super cool.*", VADER sentiment analyser would provide scores of neu: 0.32, neg: 0.0 and pos:0.68. This means that 68% of the source text is positive in nature. Additionally, the compound class calculates the sum of all lexicon ratings and normalizes the score between two extremes i.e. -1 for most negative and +1 for most positive. Henceforth, for the above-mentioned source sentence, the compound score will be 0.735 which denotes that the source sentence has a high positive sentiment. According to Hutto, a compound score of greater or equal to *0.05* would denote a positive sentiment while a compound score of less than or equal to *-0.05* would denote a negative sentiment. The neutral sentiment will be evident if the compound score lies between *-0.05 and +0.05*.

The comprehensiveness nature and a strong base of documentation gives the NLTK framework an advantage over its competitors viz. Stanford CoreNLP, SpaCy, TextBlob etc. A drawback of NLTK lies in its tokenization of sentences; the framework does not check the semantic structure and splits texts by sentences.

### 2.4.2 TextBlob

In 2014, Loria et al. [34] came up with a text processing framework named TextBlob. It is relatively lightweight Python library for sentiment analysis and natural language processing. In comparison to NLTK, TextBlob provides a better interface which is more accessible. Due to the less functional characteristics of TextBlob, it is unfair to compare TextBlob with other competitors.

However, some operations like extraction of noun phrases is easier in this framework than others. Additionally, the translate feature of TextBlob is a native wrapper around the Google Translate API.

An integrated rule-based function is present in TextBlob with dual properties i.e. subjectivity and polarity. As discussed before, the polarity score of a text sentence determines the involvement of sentiment in the sentence. For instance, the subjectivity score ranges from 0.0 to 1.0 where objectivity and subjectivity are determined as the score increases. It is to be noted that subjectivity and polarity are not correlated (Liu et al.,[31]) i.e. a sentence which is subjective in nature cannot determine any sentiment.

On the whole, TextBlob is recommended for beginners and programmers who want an easily accessible framework for sentiment analysis. However, the framework is not suitable for applications which require highly optimized performance.

### 2.4.3   SpaCy

SpaCy is an NLP framework recently developed by Mathew Honnibal (Honnibal [17]). The framework follows an approach of object-oriented programming (OOP) and is suitable for a high scaling sentiment analysis. Due to its OOP approach, the framework tokenizes the parsed text at sentence level as well as word level. The pre-built models present in SpaCy addresses important sectors like parts-of-speech (POS) tagging, classification and entity recognition.

SpaCy is one of the fastest frameworks in NLP which focuses and supports the industrial usage. The framework provides vectors for built-in words and uses a neural approach for training of models. However, it lacks flexibility when compared to NLTK and the tokenization of sentences is slower as well. In comparison to the single stemmer available in SpaCy, NLTK has nine stemmers.

NLTK is the most used NLP framework along with SpaCy. While NLTK is preferred for academia, SpaCy provides solution depending upon a problem. At the same time, TextBlob has managed to become a favourite framework for many researchers because of its ability to solve modular problems in a large pipeline. Figure 2.4 outlines the positives and negatives of NLTP and SpaCy based on Bobriakov's analysis of top NLP frameworks.

The requirements for the sentiment analysis phases of this work are simple in nature and does not require complex frameworks. Due to the standard use and functionality offered by the NLTK libraries, we have chosen NLTK framework to perform sentiment analysis. Interestingly, the NLTK library provides BLEU scores as well which is an important evaluation metric for the translation phase of this thesis.

| ⊕ PROS | ⊖ CONS |
|---|---|
| + The most well-known and full NLP library<br><br>+ Many third-party extensions<br><br>+ Plenty of approaches to each NLP task<br><br>+ Fast sentence tokenization<br><br>+ Supports the largest number of languages compared to other libraries | – Complicated to learn and use<br><br>– Quite slow<br><br>– In sentence tokenization, NLTK only splits text by sentences, without analyzing the semantic structure<br><br>– Processes strings which is not very typical for object-oriented language Python<br><br>– Doesn't provide neural network models<br><br>– No integrated word vectors |
| + The fastest NLP framework<br><br>+ Easy to learn and use because it has one single highly optimized tool for each task<br><br>+ Processes objects; more object-oriented, comparing to other libs<br><br>+ Uses neural networks for training some models<br><br>+ Provides built-in word vectors<br><br>+ Active support and development | – Lacks flexibility, comparing to NLTK<br><br>– Sentence tokenization is slower than in NLTK<br><br>– Doesn't support many languages. There are models only for 7 languages and "multi-language" models |

Natural Language ToolKit 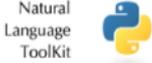

spaCy

Figure 2.4: A Comparison of NLTK and SpaCy Frameworks (Bobriakov, 2018)

## 2.5    Related Work

Several literary works of eminent authors have been studied for the sole purpose of understanding the research area of this thesis. From these learnings, some trends and approaches of such works have been identified. This section discusses the works that have taken place in the field of audio-visual communication application, sentiment analysis and language translation.

There are varied researches taking place on video analysis and extraction techniques to create audio-visual datasets. In most of these analyses, a model is trained on a large visual corpus (Kossaifi et al., [27]) in order to meet different research aims viz. exploring facial emotions (Li et al.,[30]), forensics in Virtual Reality (VR) applications (Yarramreddy et al.,[63]), emotion recognition etc. Recently, Kim et al. [24] found out that the smile intensities of a listener in a video conferencing session varies significantly if the speaker is clear to understand. Their work compared the responses of Automatic Speech Recognition Systems (ASR) present online and non-verbal speeches with the speeches which are not intelligent in nature. The analysis and research of video conferencing sessions is limited to high-tech industry members, viz. Microsoft (McDowell et al., [30]), Mitsubishi (Tian et al., [55]), Apple (Lee et al., [29]), IBM (Kennedy et al., [23]) etc and prominent figures from academia. Some researchers have been successful in inventing methods for video analysis (Walsh et al., [57], Yee et al., [64]).

In the past, researches performed sentiment analysis on a varied media sources viz. audio, visual, text to generate many corpora for research purposes. In 2016, Navas et al. [40] used the gaussian mixture modelling to generate a corpus for emotion analysis for text-to-speech functionality. In 2013, Kaushik et al. [22] used maximum entropy modelling to extract sentiments from audio streams. Since deriving sentiments from texts was common in nature, the researchers challenged themselves by exploring the concept of sentiment extraction from audio and visual corpora. Poria et al. [45] fused several clues from audio, visual and text modes to perform sentiment analysis. In the same year, Poria et al. [46] used a CNN approach to perform sentiment and emotion analysis from multimodal content. This nature of exploration from multimodal content can be seen in the works of Zadeh et al. [65] who used a neural network approach. Although Wöllmer et al. [59] took the initiative of performing sentiment analysis in a surrounding of audio-visual, there were limited innovation in this area.

It is to be noted that in most of the studies, the authors tried to run analyses using different methodologies although the surroundings of their works were in the same area. In most of the cases, several models were developed and trained on large corpora. The most common source of videos in such corpora were YouTube[12] and ExpoTV[13]. For emotion analysis, Database for Emotion Speech (DBES) in many works. All such notable works used accuracy as a standard metric to evaluate their sentiment model. In addition to or as an alternative measure in some cases, F1 scores were used. For gaussian mixture models, Navas et al. [40] used a confusion matrix to evaluate the results.

As mentioned above, since the analysis on real-time video conferencing sessions were limited, research works were read separately for video and sentiment analysis to understand the minute concepts. The whole idea is to analyse the speeches from video sessions and perform sentiment analysis based on bimodal content i.e. polarity and loudness of voice. Table 2.1 outlines the work of several authors who used different approaches to perform sentiment or emotion analysis. A significant portion of this work involves translation of speeches acquired from a video session. Thus, it is imperative to study notable works involving speech translation from one language to another.

In 2013, Krishnamoorthy et al. (2013) generated descriptions from videos using a text-mined knowledge approach. Their model was trained on a YouTube dataset which contained 2089 video segments and 85,500 sentences. In a similar context, Fernández-Torné et al. [13] worked on video descriptions of films which are dubbed into Catalan language. However, since this work was more of a study, the authors applied statistical modelling in their investigation to obtain results. A year later, Pan et al. [42] used a neural approach for the purpose of jointly translating and embedding audio-visuals and managed to bridge the gap between video and natural languages. It was around this time, the researchers adopted the neural approach to translate languages and generate text from videos due to the then disruption of NMT. Liu et al. [32] used RNN to classify texts while Venugopalan et al. [56] used the RNN-LSTM approach for a sequence-to-sequence conversion of video to text. In 2018, Afouras et al. [1] came up with the concept of speech recognition from audio-visuals using a deep learning approach. Table 2.2 outlines the work of several authors on language translation.

For most of the works, the authors used several corpora from WMT[14] for translation of languages. In addition to these, Microsoft Video Description[15](MSVD) and corpora from YouTube in cases where the model had to train videos. For works on text classifications and analysis, the model learned vocabularies from corpora like SST1, SST2[16], IMDB, etc. In contrast, Torné prepared survey to gather answers for to their research questions.

---

For evaluation, the authors focused on the correctness of translated texts from a source language. As a standard measure, the researchers used BLEU (Papineni et al., [43]) score to evaluate their works. In addition to BLEU, METEOR (Lavie et al., [28]) score was used in various cases. Metric scores like accuracy, precision, recall and F1 were used as well in some cases.

The works on language translation has been able to provide effective insights and helped to create a roadmap for the language translation module of this thesis. Additionally, BLEU can be a good measure to evaluate the correctness of translated sentences in our work.

Table 2.1: The table presents a count of studies which have combined analysis of sentiment and emotion analysis of speeches and videos. All studies use a neural network approach for their modelling except those noted (*) which include Naïve Bayes, Support Vector Machines, Gaussian Mixture Models, Maximum Entropy Modelling etc. respectively to construct the features viz. audio, visual, textual

| Author(s) | Audio-Video Type | Source | Items |
|---|---|---|---|
| Wöllmer et al.* [59] | Movie Review Video | YouTube, ExpoTV | 370 review videos |
| Poria et al.*[45] | Miscellaneous Videos | YouTube | 47 videos |
| Navas et al. [40] | Miscellaneous Speech Clips | Belfast database, VERIVOX, RUSLANA, SES database, MediaTeam Finnish Corpus, etc. | - |
| Atassi et al. [2] | Emotional Speech | DBES | 800 sentences |
| Zadeh et al.* [65] | Movie Reviews | CMU-MOSI, YouTube | 2199 utterances |
| Rosas et al. [47] | Spanish Videos | YouTube | 105 videos |
| Rosas et al. [47] | Phone Reviews | ExpoTV | 37 reviews |
| Kaushik et al. [22] | Review Comments | Amazon Product Reviews, Pros & Cons Database, Comparative Set Database | 800,000 reviews |
| Kaushik et al. [22] | Audio Reviews | YouTube | 28 videos |
| Poria et al.* [46] | Opinion Utterances | MOUD, USC IEMOCAP | 448 utterances, 5479 videos |

Table 2.2: The table presents a count of studies which have combined functionalities of audio, visual and textual translation. All studies use a neural network approach for their modelling except those noted (*) which include Text Mined Knowledge, Statistical Modelling, Unsupervised Term Discovery and Gaussian Mixture Model respectively to construct different clues viz. audio, visual, textual.

| Author(s) | Functionality | Dataset Source | Items |
|---|---|---|---|
| Venugopal et al. [56] | Video Transcription | MSVD Dataset | 120,000 sentences<br>2,000 videos |
| N. Krishnamoorthy [26] al.* (2013) | Video Description | YouTube | 85,500 sentences<br>2089 video segments |
| Liu et al. [32] | Text Classification | SST1, SST2, SUBJ, IMDB | 446,000 vocabularies |
| Fernández-Torné et al.* [13] | Audio Description | Research Questionnaire | - |
| Pan et al.* [42] | Video Language Translation | YouTube2Text | 1970 snippets |
| Bahdanau et al. [3] | Language Translation (Text) | WMT'14 | 348 million words |
| Bansal et al.* [4] | Language Translation | CALLHOME Spanish-English Dataset | 168,195 Spanish tokens<br>159,777 English tokens |
| Kamble et al.* [21] | Speech Synthesis Emotion Detection | Emotional Speech Corpus | - |
| Harwath et al. [15] | Image and Speech Learning | AMT, MSCOCO | 2.58 million images, captions |
| Afouras et al.* [1] | Neural transcription for lip reading | LR-BBC2 | 8 million utterances, 26 million-word instances |

## 2.6    Conclusion

This chapter discussed the various State of Art works in audio-visual communication systems, language processing and sentiment analysis. In Section 2.2, a background of the works in language processing have been briefly discussed. The whole idea was to give the reader an impression of the transformation of machine translation systems. An overview of various neural translation systems like GNMT, OpenNMT and Nematus was given.

In Section 2.3, an account of various APIs which support audio-visual communication systems namely, PubNub, Sinch and Tokbox, was given. Since our thesis involves analysis and translation of video chat speeches, there was need to develop an audio-visual communication system to gather the data. Section 2.4 provides insights of different frameworks which can be required for sentiment analysis. For our work, we extract speeches from our developed video chat applications, convert the speeches to text using Google's Cloud Speech-to-Text API and conduct analysis based on polarity scores and loudness of voices.

In Section 2.5, a literary review of the related works has been provided. We compared our research goals with different works and came to a conclusion to use BLEU as a standard evaluation metric for language translation model. For sentiment analysis and text analysis purposes, we have chosen the NLTK framework as it provides several functionalities. Finally, for video chat application, we have chosen to work with TokBox API as it complements the WebRTC and provide various services.

A lot of factors have decided the course of this thesis. In the next chapter, we will discuss the design and methodology of this piece of work.

# 3

# Design and Methodology

## 3.1 Introduction

Chapter Three discusses the design and methodology followed for this work. We have tried to build on the foundations of some works which have been reviewed in Chapter Two.

Firstly, this chapter will provide an overview of the system and discuss each phase in detail. There are three phases involved in this work, namely – an audio-visual communication system, analysis on speeches of actors and a neural translation system to convert English sentences to French. Secondly, the process of data gathering and transformation of speeches to text have been explained. The following section provided a detailed account of the system components. Finally, all the methodologies are integrated and a final system design is presented to the reader. Additionally, a justification on the failure of previously design approaches have been discussed.

## 3.2 System Overview

In the previous sections, a brief idea regarding the phases of this work has been given. The implementation consists of three main components: an audio-visual communication system, a speech analysis model and a neural translation model. The phases of the design process have been illustrated in Figure 3.1.

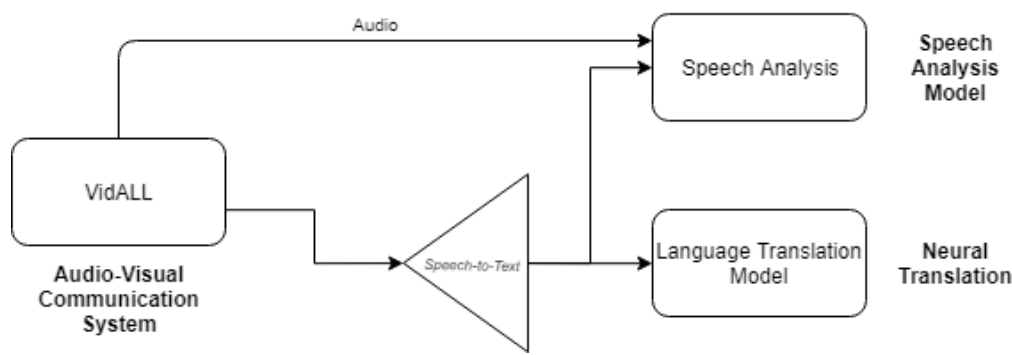

Figure 3.1: Phases of Design Process

From now on, we will refer to the audio-visual communication system as VidALL, which stands for Video-for-ALL languages. VidALL has been developed on android platform using *Java*. In VidALL, TokBox's OpenTok platform has been configured to support video streaming between the clients. For server side, we chose Heroku to set up a simple REST server.

The speeches of speakers using VidALL were extracted and passed through Google's Cloud Speech-to-Text API. Since one phase of our thesis involves language translation, it was important to derive the transcription of the speeches. English and French were chosen as the source and target language. The translation followed a neural approach where the English sentences passed through a sequential neural network to generate French translations.

For the speech analysis model, we determine the sentiment present in the speech of speaker at the beginning of his or her speech. The analysis model is based on two aspect: polarity and loudness of voice. Since the polarity score can be obtained from textual data, NLTK framework was used. Hence, the analysis model is bimodal in nature due to use of audio and textual forms.

*Python* language offers a variety of libraries which can be utilized to perform operations solving modern-day problems. Henceforth, the translation and analysis models have been coded in *Python*.

## 3.3   VidALL: Audio-Visual Communication System

The first phase involves a creation of an audio-visual communication system i.e. VidALL, as per our research objective: *to create an audio-visual communication system which has the ability to record the speech of a speaker.*

### 3.3.1   Terminologies

The OpenTok is a video platform of TokBox which supports real-time video calling between two clients. This is possible due to the presence of WebRTC in OpenTok which supports audio-visual communications. The OpenTok platform has libraries for Android, Windows, iOS etc. which can be utilized by client-side for development. Additionally, it offers a REST API.

There are mainly two components required for audio-video communication – server and client. Here, the client-side is VidALL, an android application which has been developed using OpenTok's client-side libraries. The server-side has been taken care by Heroku which uses REST services. A session connects all the clients with each other so that they can interact. Additionally, it posts events to the clients. The different terminologies within the audio-visual communication have been explained.

1. Session

   A session may be defined as a group of interactions between clients in real-time. Each session is defined by a *session ID* which is distinct in nature. Generally, the clients connect to the session after obtaining *session ID* and *token* from server.

2. Token

   A distinct authentication key which permits a client to join a session is called a *token*. The token is generated by the server whenever a client tends to join a session. While a token can expire, a session cannot expire. In VidALL, there is a role of a publisher i.e. speaker and a subscriber i.e. listener. Henceforth, a single token can be assigned different roles i.e. publisher or a subscriber, which essentially determines the authentications of the client.

3. Client

   A client may be defined as the workstation which obtains information from a server. In our case, VidALL is the client.

4. Server

   A server maybe defined as an entity which manages access to various resources in a network. In our case, Heroku used Amazon S3 as the server.

5. Connection

   A connection may be defined as the signalling association between a client and a server. A client with a *session id* and *token* after authentication will be able to establish a connection to the server. Each client connection is usually associated with a distinct *connection id*.

6. Stream

   An audio-video signal containing speaker's camera and microphone feed is called a stream. In a session, the streams are published by clients to the session.

7. Publish

   Audio-video streams are published by clients when they connect to a session. The token role of a client is publisher in this case and determines if the client can result in publishing to the session.

8. Subscribe

   Once a session is created, a client can subscribe to the streams of other clients. The token role of the other client can be of a subscriber. It is to be noted that there can be many subscribers whereas publisher can be one or two.

9. Events

   Event is referred to as the activity posted by the session when a client set up a connection.

### 3.3.2 Client-Server Architecture

VidALL retrieves the session id and token from server thereby connecting to the session using unique tokens. It is responsible for publishing and subscribing to the audio-visual streams and listening for session events. Evidently, the role of the server is to generate sessions in OpenTok cloud and tokens for clients. A client-server architecture of VidALL is illustrated in Figure 3.2.

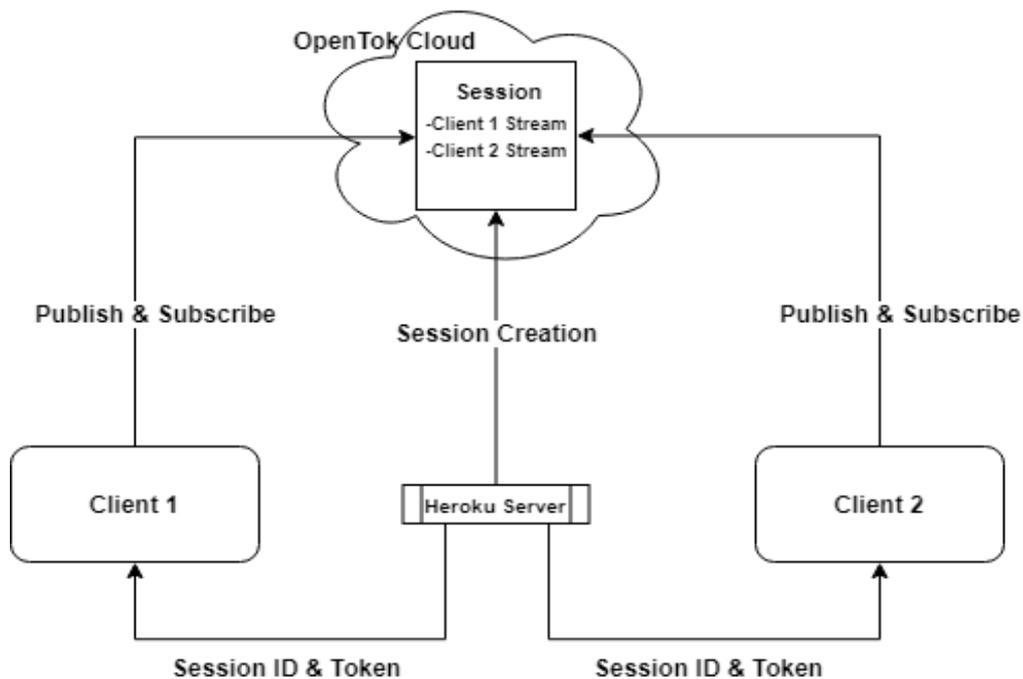

Figure 3.2: Client-Server Architecture of VidALL

This section presents a break-down of different steps taken by two clients using VidALL to set a video communication. A brief of the steps are as follows:

Step 1:

At first, a session is created by the Heroku server in the cloud through the REST API of OpenTok. Although the session is unoccupied initially, all the video communications will take place in this area.

Step 2:

Following *Step 1*, the client-side retrieves necessary session information from the server-side when a user loads VidALL. This information generally includes a unique token created by the server for authentication.

Step 3:

Following *Step 2*, the *session id* and *token* are used by the client-side to set a connection with session. Henceforth, VidALL is able to publish audio and video streams to the session and waits for a new user who join the session.

Step 4:

Following *Step 3*, another user using VidALL join the session. This event takes place once the new client receives the *session id* and *token* from the server.

Step 5:

Following *Step 4*, first client can subscribe to the stream of the second client once both the clients are connected to the session. Simultaneously, the second client can subscribe to the audio-video stream published by the first client. Thus, the audio-visual communication is established and users at both ends can communicate.

### 3.3.3   User-Interface Design

The user-interface design of VidALL is given in Figure 3.3.

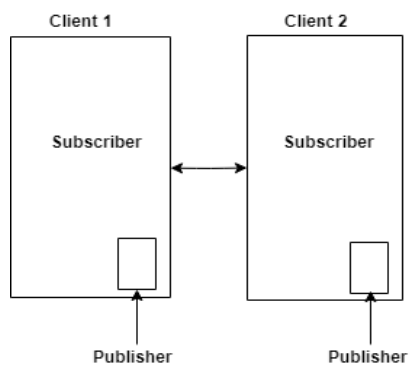

Figure 3.3:  User-Interface Design of VidALL

## 3.4    Speech Analysis Model

This phase involves creation of a model which analyses speeches obtained from VidALL. At first, video sessions of speakers are obtained from VidALL and audio is extracted using a *Python* script. To obtain transcriptions, the audios are passed through Google's Cloud Speech-to-Text API. The speech analysis model generates results based on loudness of speaker voices i.e. sound intensity and polarity scores.

Thus, this model helps to achieve two research objectives:

- *to create a small speech corpus from the audio-visual communication system*
- *to understand the sentiment of speakers in a video call*

An illustration of the speech analysis model is given in Figure 3.4.

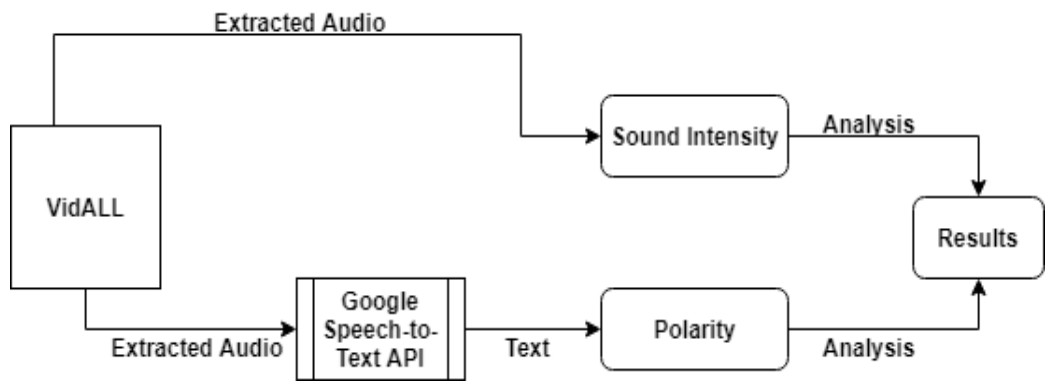

Figure 3.4:  Speech Analysis Model

The break-down of different steps involved in this model is provided below:

Step 1:

At first, the video sessions are obtained from VidALL. Since Heroku runs encryption algorithm to dynamically change web addresses of storage locations, web scrapping could not be possible. All the videos have been downloaded manually.

Step 2:

Following *Step 1*, audio files have been extracted from the audio file using a simple *Python* script. It is to be noted that the audio files are obtained in *.wav* format.

Step 3:

Following *Step 2*, the extracted audio is passed through Google's Cloud Speech-to-Text API to obtain transcription of the audio files.

Step 4:

Following *Step 2 and 3*, the intensity of speeches is measured from the extracted audio files. In parallel to this, the polarity scores of speech transcriptions are measured as well.

Step 5:

Following *Step 4*, an overall analysis of the speeches is done to understand the nature of a speaker when he or she starts a video call.

It is to be noted that the speeches from VidALL come out as audio-visuals in *MP4* format. Since our analysis require audio files, we extract the audio from the *MP4* file by running a simple *python* script. It is ensured that the format of the audio files is *.wav* in nature as Google's Cloud Speech-to-Text API takes in *.wav* audios in order to convert them into its textual form.

## 3.5 Neural Translation Model

### 3.5.1 Introduction

This section presents an overview of the neural translation model that has been developed to translate English languages to French, as per one of our research objectives. Initially, the corpora on which the model trained is described. We followed two neural models to check with model provides better results. While one of the neural models follows a simple Recurrent Neural Network (RNN), the other model follows the concept of embedding.

### 3.5.2 Data Gathering: Workshop of Machine Translation (WMT) 2010 Corpus

The WMT 2010 corpus was proposed for the fifth workshop on Statistical Machine Translation to carry out translation tasks. The training corpus was provided by Europarl which contained 1,683,156 sentence pairs of English and French. The general statistics of the Europarl WMT 2010 corpus is provided in Table 3.1.

| Items | Description | English | French |
|-------|-------------|---------|--------|
| Sentences | *Number of sentence pairs* | 1,683,156 | 1,683,156 |
| Words | *Number of words* | 47,145,288 | 50,964,362 |
| Unique Words | *Number of distinct words* | 95,846 | 123,639 |

Table 3.1: Europarl Training Corpus

As per Table 3.1, there were 1,683,156 sentence pairs of English and French in the Europarl corpora. Evidently, it would have been difficult to train such a large dataset on a CPU environment. An alternative approach would be to train the above-mentioned corpus on a GPU environment. Taking time into consideration, we reduced the size of the large corpora to meet the requirements of this work. For the ease of the reader, let us call our machine translation model as VidALL Neural Machine Translation or VidALL NMT.

The VidALL NMT corpus has been generated by running a *Python* script. A statistic of the VidALL NMT corpus has been provided in Table 3.2. It can be seen that there are more French unique words than unique English words in the vocabulary.

| Items | Description | English | French |
|---|---|---|---|
| Sentences | *Number of sentence pairs* | 123,892 | 123,892 |
| Words | *Number of words* | 1,638,435 | 1,762,601 |
| Unique Words | *Number of distinct words* | 227 | 356 |

Table 3.2: VidALL Neural Machine Translation Corpus

The VidALL NMT runs on a sequence-to-sequence RNN on TensorFlow-CPU with Keras as frontend. Keras contain a number of libraries which support the RNN modelling. Since our model runs on training samples and validates on test samples, there was no need to create a different test corpus. We have ensured that ratio of train and test samples in 80:20. Pareto's 80:20 rule states that 20% efforts can effectively yield 80% results. This principle has been followed several e-learning systems (Oztekin et al.,2013, Jin et al., 2008, and Murphy,2012). Henceforth, in this model we will use this train/test ratio.

### 3.5.3   System Overview

TensorFlow has been used for backend while Keras represents the front-end in this work. TensorFlow can be used on two modes: CPU and GPU. Given the scope of this work and related requirements, TensorFlow CPU has been used. Keras is built on TensorFlow and helps to build and validate a neural network with minimal effort. Thus, it is user-friendly due to its libraries. Additionally, the Model and Sequential APIs of Keras ensures a rapid prototyping. On the whole, the functionalities offered by TensorFlow and Keras and their flexibilities have been considered for our work.

From the literary works (Chapter 2), we found out that word of a sentence depends on the surroundings. Convolutional Neural Network (CNN) ensures independency and are widely used for pictures. Since our work is based on textual forms, RNN has been used for encoding and decoding as they help to remember the earlier results i.e. the previous words, for our case. This can be considered as a reason why RNNs are used everywhere in machine language translations.

Cho et al. (2014) introduced the concept of Gated Recurrent Unit (GRU) in 2014. Like the existing Long Short-Term Memory, GRU captures long-term dependencies as well. However, GRU uses one gate less than LSTM which makes each recurring unit to capture dependencies in an adaptive manner. Additionally, GRU consumes less memory in comparison to LSTM. In our work, GRU has been used.

Additionally, it is important to continuously update the weights of network which are involved in data training. In 2015, Kingma et al. (2015) introduced the Adam's Optimizer algorithm. In comparison to other optimizers viz. gradient descent, Adam's optimizer requires less memory, possesses intuitive interpretation capability and is computationally efficient in nature. Since our work require large terms of data to train the model, Adam's Optimizer has been used with a learning rate of 0.001, an optimum rate as proposed by Smith (2017).

In our work, we use the Sequential model from Keras to build the RNN. For the output layer, Softmax activation has been used. In Table 3.3, the specifications of the model used in this work have been outlined.

| Neural Network | *Type of Neural Network* | RNN |
|---|---|---|
| Model | *Type of Model* | Sequential |
| Learning Rate | *Rate of Learning* | 0.001 |
| Optimizer | *Optimizing Algorithm* | Adam |

Table 3.3: Language Translation Model Specifications

In this work, two RNN based model were developed and ran on the VidALL NMT corpus to achieve the results. The following models have been built and their translation results have been compared in Chapter 5 i.e. Evaluation:

- A simple RNN model
- RNN model with embedding

The next two sections present a basic architecture of the above-mentioned RNN models.

### 3.5.4   Simple RNN-based Model Architecture

This section describes the architecture of a simple RNN-based model. To ensure the model is sequence-to-sequence in nature, a simple RNN is a good representation for a sequence data. Initially, all the unique words from the VidALL NMT corpus are assigned a sequential value. In case, a word is absent in the vocabulary, it is tokenized as *<UNK>*. This is due to the fact that the words are cleaned, pre-processed, tokenized, and padded before they enter the RNN. Figure 3.5 shows the architecture of a simple RNN based model.

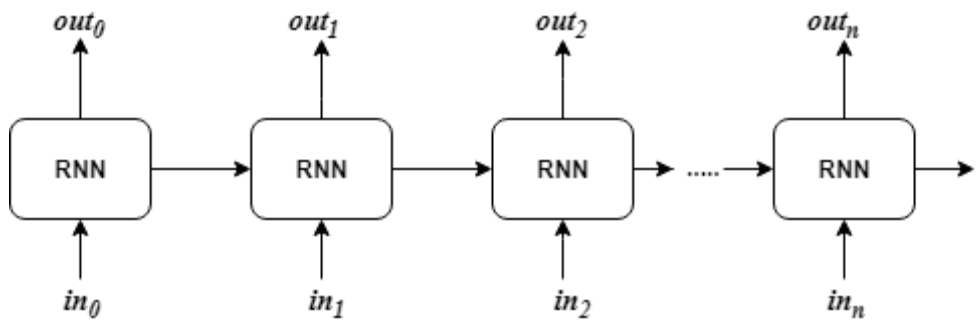

Figure 3.5:  Simple RNN-based Model Architecture

The words are obtained from a source sentence and a sequence value is assigned. From Figure 3.5, it can be seen that $in_0$, $in_1$, ..., $in_n$ are fed into the RNN which maps the training data and produces output in form of $out_0$, $out_1$, ..., $out_n$. The RNN runs a scan from left to right so that each word is dependent on the previous results. For instance, for a prediction of $out_4$, the latter gets information from $in_1$, $in_2$, and $in_3$ as well. Although, this a good characteristic of the model, it does not take care of the words which come afterwards.

As mentioned above, <UNK> is used for words which are not present in the VidALL NMT corpus. In cases, when the length of the source sentence is more than that of the target sentences, *<PAD>* can be used. For cases where a sentence reaches the end-of-line, *<EOS>* token can be used in the pre-processing of the model.

In next Chapter, we will discuss the working of this model with an example. Evidently, the model will be evaluated and results will be discussed as well.

### 3.5.5    Embedding-based RNN Model Architecture

This section describes the architecture of an embedding-based RNN model. In the earlier model, the words were tokenized in vector forms thereby ensuring a bag-of-word approach. The concept of embedding is introduced as this would help the dense vectors to represent words. In such a case, each vector represents the word projection into a continuous space of vector. It is to be noted that the word position in the vector space is learned from the text data. On the basis of the neighbouring words, the word position is learned as well. This determination of the word position in such a vector space is known as embedding. Since this work uses Keras, the embedding layer can be added easily into the RNN model.

For the output layer in this model, softmax function is used. This is due to the fact that softmax function ensures result of each unit falls in the range of *0 and 1*. In addition to this, the output of each unit is divided in such a way by the softmax function that they add up to *1*. It is this characteristic which give softmax function an edge over sigmoid function.

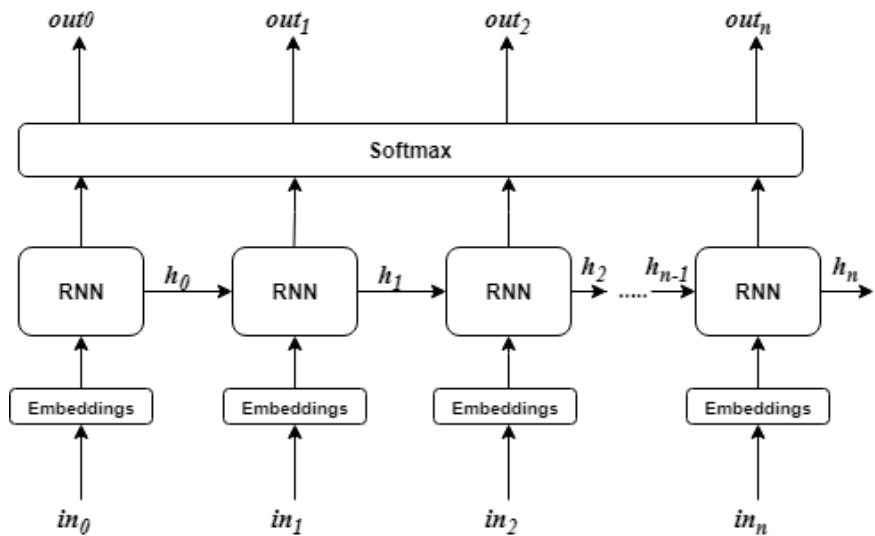

Figure 3.6:  Embedding-based RNN Model Architecture

The words are obtained from a source sentence and a sequence value is assigned. From Figure 3.6, it can be seen that $in_0$, $in_1$, …, $in_n$ are fed into the RNN which maps the training data and produces output in form of $out_0$, $out_1$, …, $out_n$. At first, the input words are embedded into vectors and fed into the RNN. The sequence of words is encoded using a GRU encoding.

The encoded words then pass through a dense layer of softmax. Eventually, the outputs are generated in the form of $out_0$, $out_1$, ..., $out_n$ in the target sentence. The functionality of softmax function on a vector is shown in the equation below. It can be seen that each unit is converted in such a way that they add up to *1*.

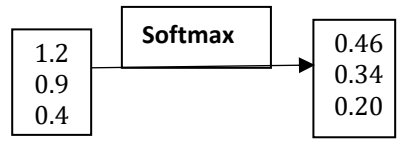

The working of the embedded-RNN model is explained along with an example in the next Chapter. Evidently, the model will be evaluated and results will be discussed as well.

## 3.6    Integration of Phases

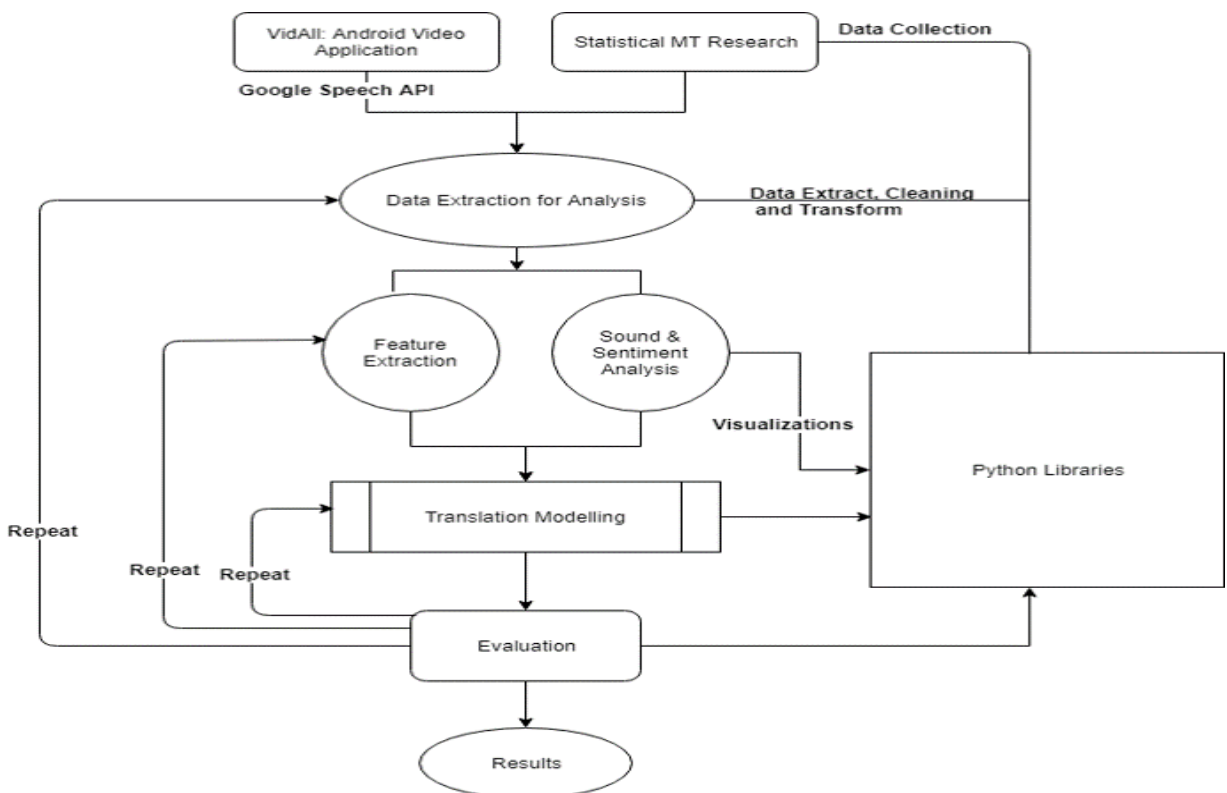

Figure 3.7:  Phase Integration

Figure 3.7 presents an illustration of the integration of all the three phases namely: development of VidALL, speech analysis model and the VidALL NMT. The repeat indicates that the work followed an agile approach during the course of the dissertation.

## 3.7 Previous Designs

In due course of this work, a lot of ideas were chanced upon to meet the research objectives. While some strategies could not be implemented, there were some which were discarded. This section discusses the designs that were previously ideated and provides a brief account of the circumstances which led to the final design of this work.

- The influence to work on speech translation and natural language processing came from Rob Dachowski's work (Dachowski, 2018). He created a model which can translate video speeches to localized languages along with translated transcriptions. Dachowski used AWS Transcribe to generate transcriptions, AWS Translate and AWS Polly to translate text and voices respectively in target languages. While the author used AWS APIs and python scripts in an AWS CLI environment, the output of this experiment came in form of a translated video. We tried to use this approach on VidALL. However, the android application could not adhere to the AWS Command Line Interface (CLI) environment. An alternative approach could have been to use the AWS services to create real-time voice translations for video communication on a web application. Recently, Tomasz Stachlewski (2019) created a web application which can record the voice of a user and translate the same into a selected target language.

- As per an early translation approach, we wanted to create real-time translated subtitles for VidALL. In order to ensure the working of translated subtitles, a simple android application was created where a speaker would be able to hear a recorded voice in *Bengali* if he or she spoke in English. A simple *Speech Recognizer* functionality was used for this purpose. However, this could not be integrated with VidALL. Since VidALL uses the services of OpenTok cloud platform, the latter prevents any other third party to use the same source of voice. For example, if a user spoke in English, the voice would directly go to the OpenTok platform. Thus, this would prevent the speech to reach to the in-built *Speech Recognizer* function of the android application. Hence, the method had to be scrapped.

- There was another attempt to create real-time translated transcriptions with VidALL by integrating the services of Google's Cloud Speech-to-Text and Translate APIs. However, speech analysis could not be possible with this approach. Moreover, Google Translate had limitations in terms of generating large chunks of outputs within a short time and the API could not be integrated with VidALL unlike Cloud Speech-to-Text API.

- Due to various literature readings, it was found that the sessions from VidALL can used for speech analysis in order to determine the nature of the speaker at the beginning of a chat session. Subsequently, a neural model was conceptualised based on state-of-art works so that the same could be integrated with VidALL at some point. However, integration of the model results with VidALL was out of scope of this thesis and required more time. Additionally, the approach could have resulted in deviation from the research objectives.

On a different note, the novel factor of this work is speech analysis and language translation from audio-visual communication sessions.

## 3.8    Summary

This chapter presented a design overview and methodologies that are involved in this piece of work. In Section 3.2, the phases of this dissertation have been briefly explained to the reader. This was followed by a detailed explanation of the three phases, namely – VidALL application development, speech analysis and VidALL NMT in sections 3.3, 3.4 and 3.5 respectively.

In a sub-section of 3.5, the data gathering process for the purpose of training VidALL NMT has been explained. Additionally, a justification was given behind the creation of VidALL NMT Corpus. The architectures of the two VidALL NMT models i.e. simple-RNN based and embedded-RNN based have also been explained. In addition to these, proper reasoning was given behind the choice of different RNN parameters.

Section 3.6 gave the reader an idea of how the dissertation would look like if all the phases are integrated. Evidently, this section illustrates the overall workflow of this dissertation. In Section 3.7, the various design strategies that were ideated or partly implemented have been discussed. Additionally, a proper justification for their failures were given. The illustrations of some early works have been provided in the Appendix section.

In next chapter, the implementation of different phases of this thesis has been explained along with results.

# 4

# Implementation

## 4.1    Introduction

Chapter Four discusses the implementation strategies followed for this piece of work. The methodologies and related designs have been explained in great detail in Chapter Three. This chapter builds on the methodologies discussed previously and provides an insight to several functionalities of this work. The findings have been presented in form of tables and visualizations.

Firstly, the chapter discusses the VidALL android application in Section 4.2. Section 4.3 provides an account of the speech analysis model which is fed by the audio files from VidALL session. In Section 4.4, the language translation model has been discussed. visualizations.

The Chapter ends with a conclusion in Section 4.5 where all the implementation work has been summarized.

## 4.2    VidALL: Audio-Visual Communication System

VidALL is an android application which makes audio-visual communication possible for two users. A unique feature of VidALL is that it allows the speaker to record his or her audio-visual during a live video calling session. A speaker can then save the recorded session to an Amazon S3 bucket. This service is provided by the Heroku which supports the server-side for this application. VidALL has been developed on android platform using Java.

There are mainly two components required for audio-video communication – server and client. Here, the client-side is VidALL, an android application which has been developed using OpenTok's client-side libraries. The server-side has been taken care by Heroku which uses REST services. A session connects all the clients with each other so that they can interact. Additionally, it posts events to the clients.

### 4.2.1    User-Interface

This section discusses the user-interface of the VidALL application. It is simple in nature. The outer layer is denoted by the *Vertical Layout*. In Figure 4.1, the user-interface design has been presented. As per the discussion in Chapter 3, the publisher is the space occupied by the speaker whereas the visual of the listener occupies the subscriber space.

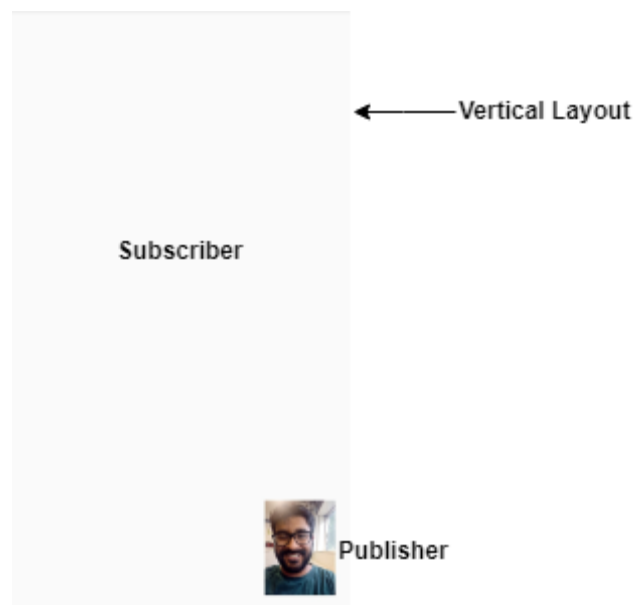

Figure 4.1 User Interface Design

### 4.2.2    Application Development

Once the user interface design is done, the OpenTok SDKs and libraries are used to build the application. At first, the necessary classes from OpenTok are loaded in the *MainActivity.java* file. In the *MainActivity* class of the file, variables to store the values of token, API key and session ID of OpenTok are added. All the necessary permissions are setup and passed in order to set up authentication and the connection to an OpenTok session is established.

Since VidALL require audio and visuals of individuals, it is important to set up permissions. The permission requests to a user to use the camera and recorder of his or her device has been implemented with the help of *EasyPermissions* library. In the running state, VidALL checks if the user has granted the necessary permissions. The requestPermissions feature of the library prompts a user for mic and camera permissions.

The Session.Builder() which creates a session object the OpenTok API key and Session ID as parameters. While the session is instantiated using the Session.Build.build() feature, the .setListener() function creates an object which implement the interface of SessionListener. The connect() method of Session accepts the token and establishes a connection to OpenTok session. The *SessionListener.onDisconnected()* is called if a client disconnects from the session. As a standard approach, *.onError()* method is called whenever any failure happens.

It is to be noted that other methods of *SessionListener()* i.e. *onStreamReceived()* and *onStreamDropped()* are responsible for publishing and stopping a stream on the OpenTok session. For publishing and subscribing audio-visuals to the session, Publisher and Subscriber class from OpenTok SDK has been used.

Finally, the *onStreamReceived* method is used when two clients are connected and audio-visuals are streamed.

### 4.2.3 Session Storage

The storing of the sessions has been achieved through the REST API of OpenTok. A session recording can be possible with at least one connected client. The session streams are archived as the client start-and-stop the publishing stream. The recorded session can then be stored by the speaker in the Amazon s3 bucket whose services are provided by Heroku. Only the speaker has access to his or her recorded message which is stored the Amazon S3 environment.

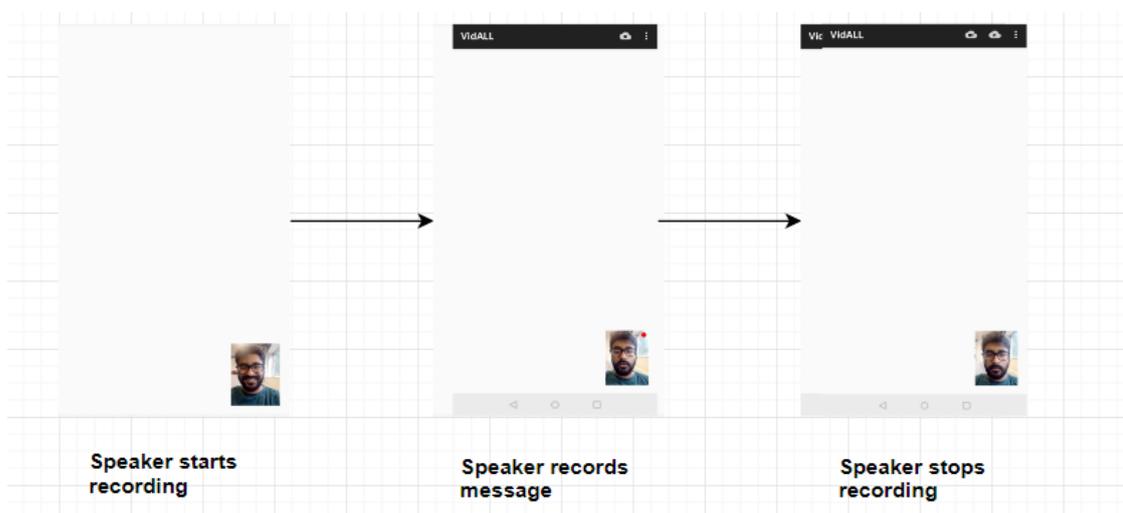

Figure 4.2: VidALL Session Recorder Phases

## 4.3 Speech Analysis

This section presents a discussion on the speeches generated from VidALL video chat sessions. The extracted videos are converted to audio files which are then analysed on the basis of their polarity scores and sound intensity. In this phase, we try to find out if polarity scores and sound intensity are correlated to each other. Additionally, the nature of individuals in the initial seconds of their video sessions is understood. This is possible due to two types of analyses: voice analysis and sentiment analysis.

### 4.3.1 Data Extraction

According to the literary works in Chapter 2, it was realized that sentiment analysis of video chat sessions is very few in number. This is due to the fact that organizations or third-party do not release video chat data in order to maintain user-privacy and user-security. For this purpose, VidALL was developed. Due to the unavailability of necessary corpuses, a session recording feature was introduced in VidALL. A user can record his or her session while talking to the other person. Such sessions are important for our work as they ensure a real-time reaction of the individual. This would help to generate authentic results.

The videos were extracted from VidALL and were converted to audio forms using a *Python* script. The subprocess library in Python offers several features to stream and extract audio and video files. Using *ffmpeg* command, the videos were converted to audio files in *.wav* format. It is to be noted that this work discarded the visual features from the sessions to ensure user security. Additionally, this was beyond the scope of our research aims.

To understand the working of VidALL on various smartphone devices, individuals were invited for tests. In due course of time, twenty audio files were obtained on the whole. Each individual was asked to start their video sessions with specific sentences. For such demonstrations, five different sentences were selected. Thus, each distinct sentence was used by two females and two males.

A statistic of the speech analysis corpus is given in Table 4.1. This meets one of our research objectives. The sentences which were given to the individuals are stated in Table 4.2

| Audio files | Female Speakers | Male Speakers | Unique Sentences |
| --- | --- | --- | --- |
| 20 | 10 | 10 | 5 |

Table 4.1: Speech Analysis Corpus Statistics

| Sentences |
| --- |
| *1. Hi, how are you?* |
| *2.You know, I am finally* |
| *feeling happy.* |
| *3. Hi, good to see you.* |
| *4. It feels great to talk to you* |
| *after such a long time.* |
| *5. Oh my god, look at you.* |

Table 4.2: Sentences Given to Speakers

## 4.3.2    Pre-processing

The audio files were cleaned and pre-processed before they were fed into the voice analysis and sentiment analysis model. As part of the pre-processing stage for voice analysis, the first five seconds of the session recordings are extracted. The approach was needed in order to determine the behaviour of an individual during the starting of a video call.

To calculate the polarity scores, NLTK library has been used. However, the library take input in textual forms. Hence, the trimmed audio files were passed through Google's Cloud Speech-to-Text API to obtain their respective transcriptions.

The pre-processing of audio files included the following steps:

- Conversion of extracted video files to audio files
- Extraction of initial *5 seconds* of audio files
- Feeding of extracted audio files into Cloud-to-Text API
- Feeding of trimmed audio files in voice analysis model
- Feeding of audio transcriptions in sentiment analysis model

## 4.3.3    Voice Analysis

Voice plays a vital role in conveying emotions. Generally, the voice of a speaker changes according to the reactions of a person to certain situations.  Thus, the loudness varies depending upon the mood of the speaker, the content of the message and the context. Hence, we try to investigate the relation of loudness and sentiment involved in a voice in this work.

Pitch may be referred to as the vibration frequency of vocal folds. In actual, the oscillation sizes of vocal folds affect the loudness (Hacki, 1996). In other words, these oscillation sizes are also known as amplitude. The loudness of a voice is an alternative measure for sound or acoustic intensity. According to Collins (2010), the sound intensity and amplitude of an audio wave are inversely proportional to each other.

In Python, Librosa (McFee et al, 2015) library was used which is responsible for analysis of speeches and signals. At first the audio files were loaded and a frame size was assigned to each file using the *librosa.stft* function. This function is responsible to compute the short-time Fourier transform (STFT) of the audio files.

For instance, let us take the example of sentence 4 from Table 4.2 i.e. "*It feels great to talk to you after such a long time.*". Figure 4.1 represents the wave plot of *sentence 4*.

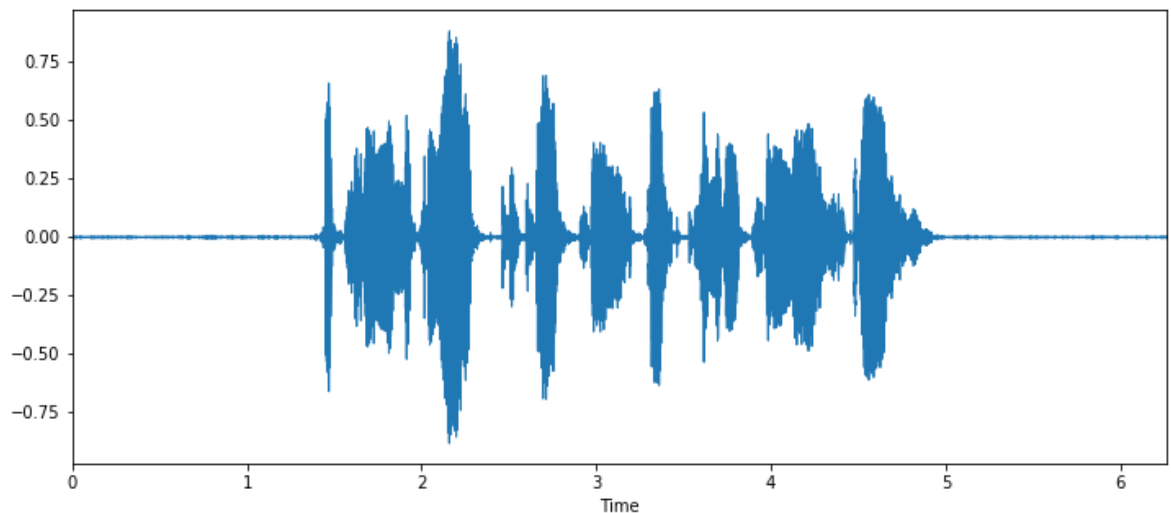

Figure 4.3: Waveform of Sentence: "*It feels great to talk to you after such a long time.*"

In Figure 4.3, it can be seen that intensity of the voice varies with every utterance. The speaker took time to start a speech and took *5 seconds* to complete *sentence 4*. The figure gives an impression that the speaker could have been calm and composed during the speech. However, the nature of the speaker can not be determined solely on the basis of time-frame. In his work, Hacki (1996) provided a range of sound intensities which determine the nature of an individual i.e. whisper, low voice, normal voice, shouting etc. Since sound intensity varies over time, a mapping of the intensities from the sample speeches with Hacki's range can determine an individual behaviour.

The general perception of voice intensity is that it is logarithmic in nature. It is important to find out the logarithmic amplitude and its relation with the sound intensity. It is to be noted that the unit of intensity is decibel(dB).

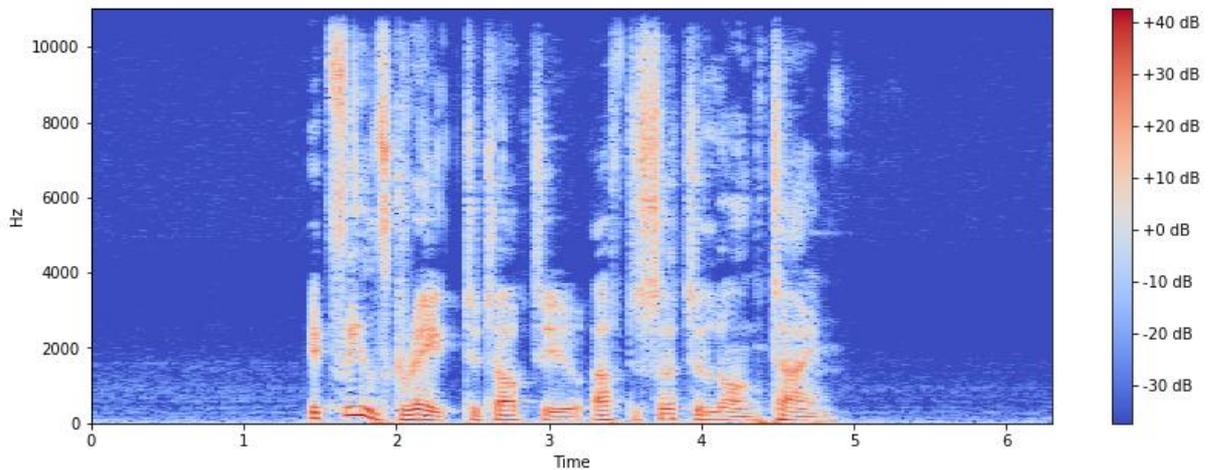

Figure 4.4: MEL-Spectrogram of Sentence: "*It feels great to talk to you after such a long time.*"

Figure 4.4 represents a MEL-Spectrogram of *Sentence 4.* An idea of the sound-intensity can be inferred from the visualization. The sound intensities can be calculated using the *librosa*. In cases where amplitude is derived, the *amplitude_to_db* function is used to determine the sound intensity. Due to the presence of different intensities, mean sound intensity was chosen as an unbiased parameter to compare with the range provided by Hacki (1996).

The vocal comparators from the works of Hacki (1996) state that a vocal sound range from 10 dB to 90 dB. The threshold of a vocal sound is around 110 dB, above which the sound yields a painful sensation.

| whisper | quiet voice | normal speech | shouting | singing | pain |
|---------|-------------|---------------|----------|---------|------|
| 10 dB | 35-40 dB | 50-70 dB | 80-90 dB | 100 dB | 110-130 dB |

Table 4.3: Hacki's Vocal Intensity Comparators (Hacki, 1996)

Table 4.3 shows the vocal intensity comparators suggested by Hacki. Since the samples of our work include normal conversations, the mean vocal intensities of the audio samples have been classified as per Table 4.3. Additionally, the vocal intensities have been normalized for our analysis.

| whisper | low voice | normal speech | excited |
|---------|-----------|---------------|---------|
| <20 dB | 20-40 dB | 40-70 dB | >70 dB |

Table 4.4: Model Vocal Intensity Comparators

The mean sound intensity for Sentence 4 was found to be *26.24 dB*. Hence, the speaker spoke in a *low voice*. The vocal intensities of all the samples were found out and classified according to the Table 4.4.

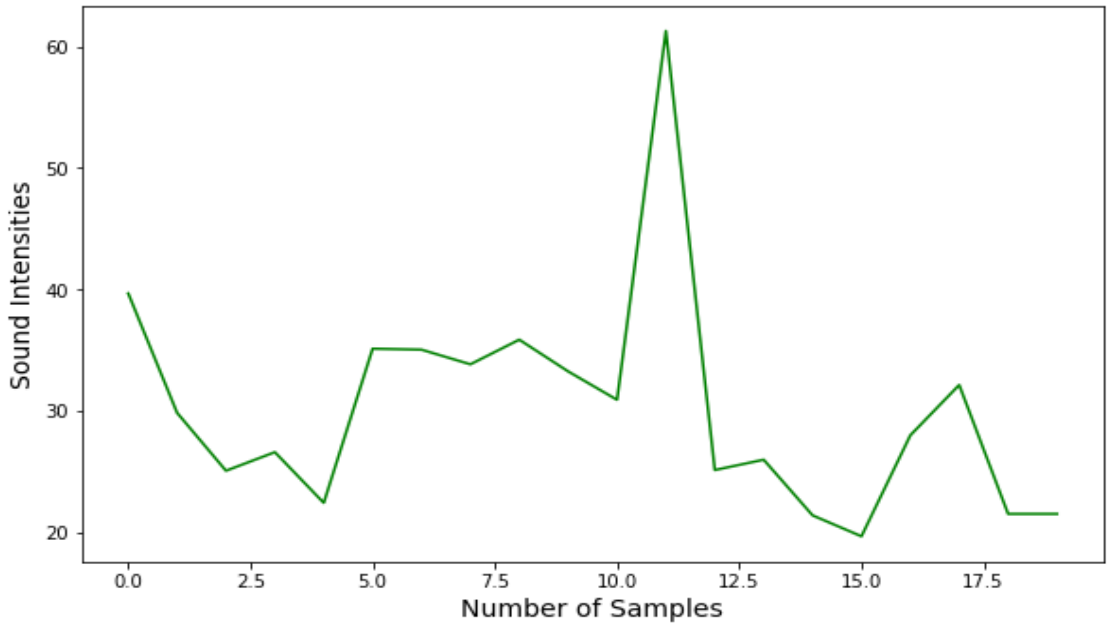

Figure 4.5: Sound Intensities in Audio Samples

From Figure 4.5 it can be deduced that the mean intensities of sample lied between 20 dB and 40 dB in most cases. One of the speakers had an acoustic intensity greater than 60 dB while another individual recorded the intensity which was lesser than 20 dB.

Additionally, analyses on Sentence 1 and Sentence 2 led to some key findings. It was observed that female speakers recorded higher vocal intensities than male intensities. Interestingly, the intensities of both male and female speaker lie in the *low voice* range. Table 4.5 accounts the intensities of female and male speakers.

| Sentences | Female Speaker | Male Speaker |
|---|---|---|
| *1. Hello, how are you?* | 39.69 dB | 34.13 dB |
| *2. You know I am finally feeling happy.* | 33.08 dB | 30.06 dB |

Table 4.5: Sound Intensities for Sentence 1 and Sentence 2

Although it has been established that sound intensity can be related to the loudness level, a sentiment analysis on the voice transcriptions is required to understand speaker behaviour.

### 4.3.4 Sentiment Analysis

Sentiment analysis is done on the audio samples to identify and understand the opinions which are expressed by the speakers. The transcript of audio samples obtained from the Cloud Speech-to-Text API are analysed one by one using NLTK's VADER sentiment analyser. As explained in Chapter 3, VADER generates the overall sentiment score of a speech transcript in form of *Compound Score*. The sentiment analyser categorizes pieces of text as neutral if the compound score lies between -0.05 to 0.05. For other cases, the analyser classifies the texts as positive and negative if the scores lie above 0.05 or below -0.05 respectively.

For our work, the classifications have been further categorized into very negative and very positive sentiment. The categorizations have been explained in Table 4.6.

| very negative | negative | neutral | positive | very positive |
|---|---|---|---|---|
| -1.0 to -0.5 | -0.5 to -0.05 | -0.05 to 0.05 | 0.05 to 0.5 | 0.5 to 1.0 |

Table 4.6: Classification of Sentiment Scores

Based on the above classifications, the transcriptions of each audio files have been assigned a sentiment score and categorized according to Table 4.7. It has been found that only Sentence *1* is neutral in nature. Interestingly, *Sentence 1* is interrogative and is widely used as a starting sentence for many conversations. At the same time, *Sentence 5* starts on an exclamatory note and could have proportions of negative sentiment present. However, its overall score is positive according to the analyser.

| Sentences | Overall Score | Category |
|---|---|---|
| *1. Hello, how are you?* | 0 | neutral |
| *2. You know I am finally feeling happy.* | 0.6369 | very positive |
| *3. Hi, good to see you.* | 0.4404 | positive |
| *4. It feels great to talk to you after such a long time.* | 0.6249 | very positive |
| *5. oh my god, look at you.* | 0.2732 | positive |

Table 4.7: Sentiment Scores for Five Sentences

Figure 4.6 illustrates the distribution of sentiments in the transcriptions of sample audio files. As discussed earlier, there is no neutral sentiment present in all the sentences. On the whole, neutral and positive sentiments form most of the distribution.

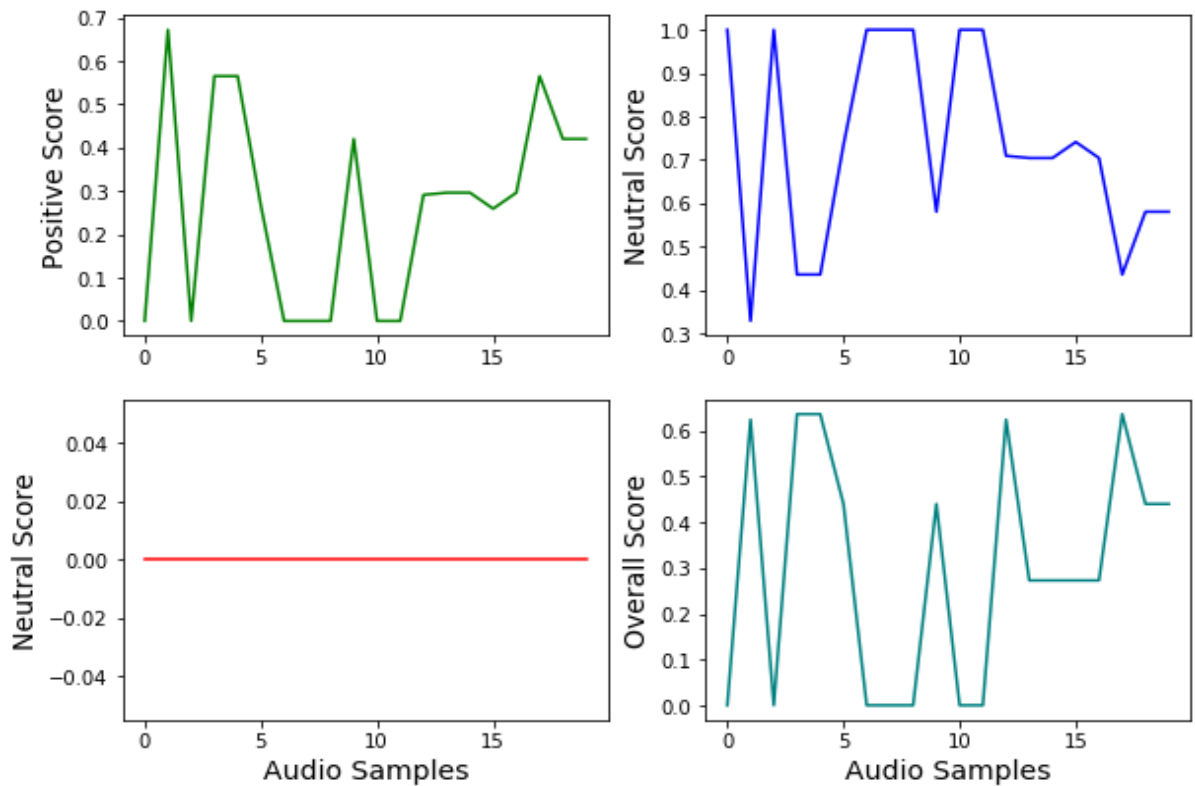

Figure 4.6: Sentiment in Audio Samples

## 4.3.5    Voice-Sentiment Analysis

This section integrates the sentiment scores and sound intensities and their respective classifications thereby providing insights. Out of the 20 audio samples, 7 are found to be neutral in nature. A visualization of positive sentiment distributions and sound intensity has been generated to understand the relation between sentiments and sound intensities. Positive sentiment has been chosen over other sentiments due to its more presence. Additionally, all the sentences were classified in the positive category.

Figure 4.7 presents the correlation of positive sentiment distribution and sound intensities on samples where positive sentiment is present. The positive scores have been converted into percentages while the sound intensities have been converted into its percentage form, taking the higher conversation limit of 90 dB as the upper limit.

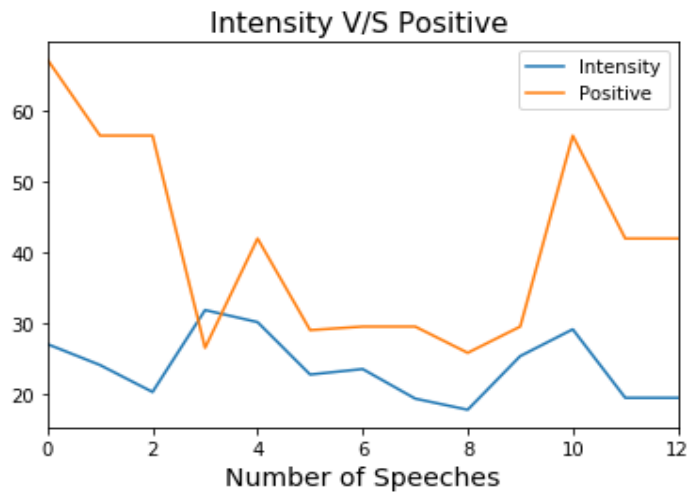

Figure 4.7: Positive Sentiments and Sound Intensities in Audio Samples

From Figure 4.7, it can be seen that fifth speech onwards the positive sentiments reacted directly to the intensity changes. However, the same can not be same for the previous speech samples. Thus, we conclude that sentiment and intensity do not have a correlation. At the same time, the best parameter to judge the behaviour of an individual in a video chat would be through sentiments.

### 4.3.6 Conclusion

The Speech Analysis on the sample audio files led to some important findings:

- Sentiment and Sound Intensity do not have a correlation
- The behaviour of an individual can be analysed by the sentiments rather than the intensity of their speeches
- Most individuals start their conversation on a positive note
- Female speakers tend to have higher speech intensities than male speakers

## 4.4 VidALL NMT: Neural Translation Model

The VidALL is a neural network-based model which translates English languages to French languages. The design and methodology of RNN models have been explained in great depth in Chapter 3. The two RNN-based models are run on TensorFlow-CPU with Keras as the front-end.

### 4.4.1 Dataset

As mentioned in Chapter 3, the VidALL NMT corpus is created for training the two RNN-based models. The corpus is derived from the Europarl Training Corpus which was used in WMT 2010. The dataset contains 123,892 sentence pairs of English and French in two different files.

At first, the French and English datasets are loaded into the system and the sentences are stored in lists, namely, *english_sentences* and *french_sentences*. It is then checked whether the sentences are pre-processed i.e. conversion of all texts to lowercase, delimitation of punctuations using spaces. A rich and complex vocabulary makes the problem more complex. Hence, initial focus has been given to the vocabularies of English and French. In total, the number of French and English words were found to be 1,762,601 and 1,638,435 respectively. Since there was a lot of repetitive words, distinct words for both English and French were determined. Along with this, the most common ten words for both English and French were obtained.

### 4.4.2 Model Pre-process

This section discusses the pre-process strategies that are followed in the run up to RNN models. It is to be noted that the text data is not inserted as input to the RNN model. Instead, the texts are converted into integer sequences using tokenization and padding methodologies. While the words are tokenized to ids, padding is applied to ensure all the sequences have equal length. Tokenization could have been applied to each letter as well. However, letter-tokenization could have made the process more complex.

Keras contain a *Tokenizer* function which help to convert each sentence into a sequence of words. The tokenization process has been explained in Figure 4.6. A *tokenize* function was created for this purpose. For example, after tokenization for every input of *a* in a sentence, the output was 3.

Keras' pad_*sequences* function has been used to ensure English and French sentences have equal length. An example for padding has been provided below.

```
              Input: [18,19,3,20,21]
              Output:[18,19,3,20,21,0,0,0,0,0]

{'the': 1, 'quick': 2, 'a': 3, 'brown': 4, 'fox': 5, 'jumps': 6, 'over'
: 7, 'lazy': 8, 'dog': 9, 'by': 10, 'jove': 11, 'my': 12, 'study': 13,
'of': 14, 'lexicography': 15, 'won': 16, 'prize': 17, 'this': 18, 'is':
19, 'short': 20, 'sentence': 21}

Sequence 1 in x
  Input:  The quick brown fox jumps over the lazy dog .
  Output: [1, 2, 4, 5, 6, 7, 1, 8, 9]
Sequence 2 in x
  Input:  By Jove , my quick study of lexicography won a prize .
  Output: [10, 11, 12, 2, 13, 14, 15, 16, 3, 17]
Sequence 3 in x
  Input:  This is a short sentence .
  Output: [18, 19, 3, 20, 21]
```

Figure 4.8: Tokenization of Sentences

In the final pre-processing pipeline, it was ensured that the labels are reshaped as the sparse categorical crossentropy function of Kera's require 3-dimensional labels.

### 4.4.3 Model 1: Simple RNN-based Model

The model takes the input ids of each word and are passed through the RNN layer. A *logits_to_text* function is used in the middle layer which bridges the gap between the French translations and the. This function helps to understand the output of the model in a better way.

Figure 4.9 shows a basic RNN-based model. The number 8 is assigned to "the" which is passed through the RNN to map with 13 which is a token assigned to "le", the French translation of "the".

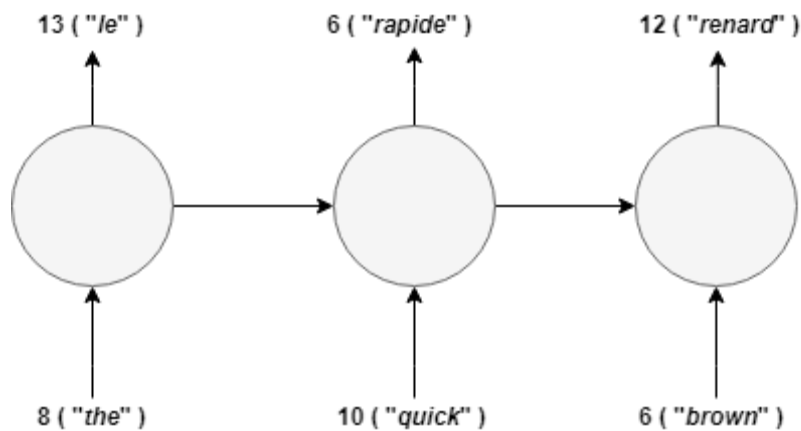

Figure 4.9: Simple RNN-based Model

| Layer (type) | Output Shape | Param # |
|---|---|---|
| input_1(InputLayer) | (None, 21, 1) | 0 |
| gru_1(GRU) | (None, 21, 64) | 12672 |
| time_distributed_1 | (None, 21, 350) | 22750 |
| activation_1 | (None, 21, 350) | 0 |

Total params:  35,422

Trainable params:  354,22

Non-Trainable params:  0

Table 4.8:  Simple RNN-based Model Summary

Table 4.8 shows the layers of the RNN model. Some characteristics of this model are mentioned below:

- The model uses *GRU* instead of a LSTM

- The time distributed layer is occupied by *logits*

- The model uses *softmax* as the activation function for the output layer

- For model compilation, *sparse categorical crossentropy* function is used as loss function

- *Adam's* Optimizer is used to optimize the model

- Learning rate = $1\times10^{-3}$

The RNN-based model was trained on 99113 samples and was validated on 24779 samples. This ensures an 80:20 training to test ratio. The model was trained over 10 epochs with a batch size of 1024.

On the whole, the model achieved an accuracy of 61.29% during the training process.

| Source Sentence | *new jersey is sometimes quiet during autumn, and it is snowy in april.* |
|---|---|
| Target Sentence | *new jersey est parfois calme pendant l' automne, et il est neigeux en avril* |
| Predicted Sentence | *new jersey est parfois en en et il est est en en* |

Table 4.9:  Simple RNN-based Model Results

From Table 4.9, it can be seen that the model could not predict an accurate result for the source sentence.

### 4.4.4    Model 2:  Embedded-RNN Model

In the embedded-RNN model the word tokens first pass through an embedding layer. This results in vector representation of any term which is closer to similar terms in a space of n-dimension. Here, n represents the embedding vector size.

The term "the" passes through an embedding layer. The vector representation of the term can be seen in Figure 4.10. In the next steps, these embeddings pass through the logits function which eventually map the token of every English word with that of the targeted French word.

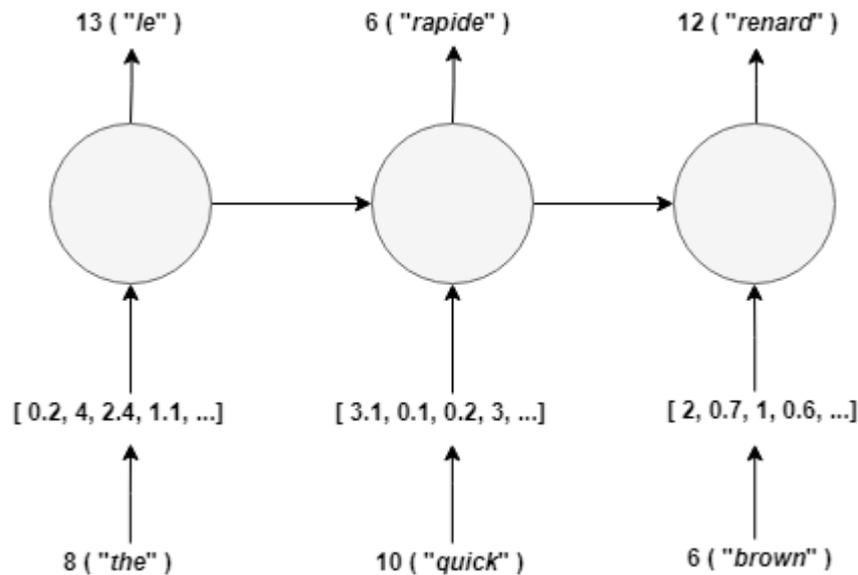

Figure 4.10:  Embedded RNN-based Model

| Layer (type) | Output Shape | Param # |
|---|---|---|
| Embedding_1(Embedding) | (None, 21, 64) | 22400 |
| gru_1(GRU) | (None, 21, 64) | 24768 |
| time_distributed_1 | (None, 21, 350) | 22750 |
| Total params:  69,918 | | |
| Trainable params:  69,918 | | |
| Non-Trainable params:  0 | | |

Table 4.10:  Embedded-based RNN Model Summary

Table 4.9 shows the layers of the embedded-RNN based model. Some characteristics of the model are mentioned below:

- The embedding layer maps every input word to a 64-dimensional vector
- GRU has been used instead of LSTM
- The time distributed layer is occupied by *logits*
- The output layer of this model is dense and fully connected. This results in probability distribution for each word using a *softmax* activation function
- For model compilation, *sparse categorical crossentropy* function is used as loss function
- *Adam's* Optimizer is used to optimize the model
- Learning rate = $1 \times 10^{-3}$

The embedded-RNN based model was trained on 99113 samples and was validated on 24779 samples. This ensures an 80:20 training to test ratio. The model was trained over 10 epochs with a batch size of 1024.

On the whole, the model achieved an accuracy of 82.71% during the training process.

| | |
|---|---|
| Source Sentence | *new jersey is sometimes quiet during autumn, and it is snowy in april.* |
| Target Sentence | *new jersey est parfois calme pendant l' automne, et il est neigeux en avril* |
| Predicted Sentence | *new jersey est parfois calmne en l' et il il est en en en* |

Table 4.11: Embedded-RNN based Model Results

From Figure 4.11, it can be seen that the prediction of the target sentence improved in this model and yielded better results than the earlier RNN-based model.

### 4.4.5   Conclusion

This section discussed on the translation of two languages using an RNN based neural approach. Some findings have been obtained which have been mentioned below:

- The model accuracy of embedded-RNN was higher than that of simple RNN-based model
- Embedded-RNN model showed better correctness of the predicted sentences than the simple RNN-based model
- The computational time of simple-RNNs were slower than the embedded-RNNs as the number of epochs increased

## 4.5   Summary

This chapter discussed the implementation techniques of the different phases of this work and the related results which followed. In 4.2, the creation process of VidALL was discussed along with challenges.

4.3 gave a brief account of the sound and sentiment analysis. This was followed by integration of the insights of both the analyses. Key conclusions were obtained where it was found out that female speakers tend to have higher vocal intensities than the male speakers. Additionally, the importance of sentiment over sound intensity as an indicator to understand human behaviour was discussed.

In 4.4, the pre-process pipelining of VidALL NMT was discussed. In addition to this, the basic RNN and embedded-RNNs model were discussed along with its characteristics. It was seen that embedded-RNNs performed better than the normal RNNs.

# 5

# Evaluation

## 5.1   Approach

Chapter Five discusses the various evaluation strategies that have been followed in this work. Since our work contains three phrases, it is important to evaluate each phase based on different parameters. For this purpose, the chapter is divided into sub-sections accordingly and an overview of the evaluation techniques in the following sections have been mentioned below:

- Firstly, Section 5.2 evaluates the security and privacy of the VidALL application. Although VidALL uses OpenTok platform to ensure audio-visual communication, a security audit is ensured. The theoretical concepts of security and privacy are mapped with the features of VidALL. The whole idea is to evaluate VidALL on the basis of its architectural design, user-security and user-privacy.

- Section 5.3 presents a case study on emotional speech dataset using our speech analysis model. The RAVDESS Emotional Corpus has been used for this purpose. The whole idea is to validate the observations derived from the speech analysis model. Since our speech analysis model analysed the speeches generated from VidALL sessions, there was a need to check whether same observations can be inferred from a larger dataset.

- Section 5.4 presents the evaluation strategy of VidALL NMT. The translation model is evaluated on the basis of its model performance and translation correctness. Additionally, a case study has been provided where the speeches generated from VidALL are fed into the VidALL NMT and the translated results are checked on the basis of its correctness.

The chapter summarizes the inferences obtained from the evaluation strategies in Section 5.5.

## 5.2    VidALL: Evaluation and Security-Privacy

As an evaluation approach for VidALL, the security and privacy considerations for the audio-visual communication system have been outlined. The related threats and mitigation process have been discussed as well.

### 5.2.1    Security Considerations and Threats

Security ensures assurance, maintain confidentiality, and blocks any unauthorized changes. Since VidALL takes the services of OpenTok API for video and audio streaming, the security of the application is dependent on the API. It is to be noted that any security failure will result in data loss or application failure. This section provides an account of the security considerations for VidALL and the possible threats the application might be subjected to.

1. Surveillance

   Surveillance is type of security threat where an attacker does not harm the user but checks the activity of the user from time to time. Such a threat can only happen if the developer shares his or her TokBox account credentials with a third-party. In such a scenario, the attacker will be able to track the log activity of VidALL.

2. Compromise of Data

   The compromise of data is a common type of security failure. OpenTok WebRTC encrypts the audio and visual data end-to-end with AES-128 while HMAC-SHA1 checks the integrity of the data. The design of VidALL is such that only a speaker will be able to record his or her video session. Thus, in a video chat session, an individual will not be able to record the session of the other person thereby ensuring data protection. However, compromise of data can happen if the speaker shares the online link of the recorded session with an unreliable third-party. Although, Heroku ensures inaccessibility of data if accessed after a prolonged duration, data compromise can happen if any third party access the data quickly.

3. Intrusion

   Any threat which interferes with the activity of application users can be termed as intrusion. Intrusion is common in nature. For VidALL, if a speaker shares his or her OpenTok account credentials with an individual who eventually tampers with the secret token, will encourage intrusion. The secret token and project id which is crucial to the configuration of the android application with OpenTok can result in denial-of-service if the project id and secret token are deleted or changed.

4. Denial-of-Service

   VidALL has been developed to ensure audio-visual communication of two devices at a time. If a third user or more try to access VidALL, it will result in a deadlock situation. Thus, a denial-of-service will take place and the application will crash. Additionally, the application will crash if the project key and API token are tampered or expires. However, this will be a failure due to the developer's negligence.

## 5.2.2    Privacy Considerations and Threats

Privacy ensures trust and confidence of an individual on any application. A privacy threat can harm an individual financially, physically, reputationally etc. This section provides an account of the privacy considerations for VidALL and the possible threats the application might be subjected to.

1. Correlation

   Correlation occurs when an attacker or third-party has access to several sensitive information or activity and correlates them. This generally includes duration of activity, traffic analysis, determining network settings and information of physical devices, etc. Such a privacy threat can only happen if the developer shares his or her TokBox credentials.

2. Identification

   Identification is closely related to Correlation. This type of privacy threat enables an attacker to identify individuals and control them. The security design of VidALL prevents identification.

3. Disclosure

   Disclosure takes place when a third-party share information of an individual without his or her authorization. Since VidALL prevents access of any third-party to user data, the possibility of any disclosure threat is minimal.

4. Secondary Use

   Secondary use refers to the event when an attacker has the credentials of any individual without his or her realization. Secondary use can harm an individual in terms of reputation. However, since VidALL does not have a user login feature, it eradicates this threat. Thus, two speakers will have access to VidALL at a single point without the involvement of any credentials. The possibility of Secondary Use threat can happen if the attacker knows the credentials of the developer's TokBox account. However, the attacker cannot cause any privacy attack to a user except indulging in the surveillance of application activity.

### 5.2.3 Threat Mitigations

VidALL has ensured steps to prevent common security threats. An account of possible threats has been discussed in the above-mentioned sections. This section discusses how the possible threats related to security and privacy can be mitigated.

1. Data Minimization

   The audio and video data in a chat session is streamed through TokBox which ensures end-to-end data encryption using cryptographic algorithms. A common security measure to prevent data loss or theft is by minimizing the quantity of data. The users in a VidALL chat session does not require credentials to use the application. This ensures user anonymity and confidentiality of identity. Data minimization from user end can result in minimization of user privacy threats.

2. User Participation

   Although the role of TokBox API ensures mitigation of security threats, user participation is also required. For instance, user can keep away unwanted listeners while engaging in a video chat session. This will prevent the users from any privacy related threats.

3. Security Measures

   The encryption measures taken by TokBox ensures authentication of peer identity and maintains confidentiality. Although the user anonymity is ensured, protocol anonymity couldn't be attained with this application. A suggestion would be use to multiple protocols which can confuse any attacker. Another security concern can be traffic analysis if an attacker has access to the developer's TokBox account.

On the whole, all the possible security and privacy threats can be mitigated if a proper security action plan is in place.

### 5.2.4 Human Evaluation

VidALL was passed through various trails and strategies involving human evaluation:

- Initially, individuals were invited to record video sessions in VidALL. While most of the trials were successful, VidALL could not run on smartphones having android version less than 24.

- In comparison to WhatsApp and Skype video applications, VidALL is slower. At times, the connection setup took more than a minute.

- Due to OpenTok's video platform support, VidALL ran properly. However, when two nearby devices were connected, a noisy echo was produced.

- The users were able to listen to each other clearly in an undisturbed environment.

## 5.3 Case Study: Speech Analysis Evaluation on RAVDESS Corpus

In Chapter 4, the speech analysis model ran on 20 samples of audio clips. Since the count for the number of audio-clips was less, it was necessary to run the model on a larger number of clips. For this purpose, the audio clips from Ryerson Audio-Visual Database of Emotional Speech and Song (RAVDESS) were chosen.

RAVDESS contains audio-visual clips of 12 female and 12 male actors. On the whole, around 7000 files are present in the corpus. Since our model is based on speech analysis of audio clips, the audio-clip segment of RAVDESS corpus was chosen to evaluate our model which contain 2880 files.

The analysis model was fed with 440 audio files which contained voices of male and female actors. The sentiment associated with these audio-clip transcriptions was mapped with the sound intensity as shown in Figure 5.1.

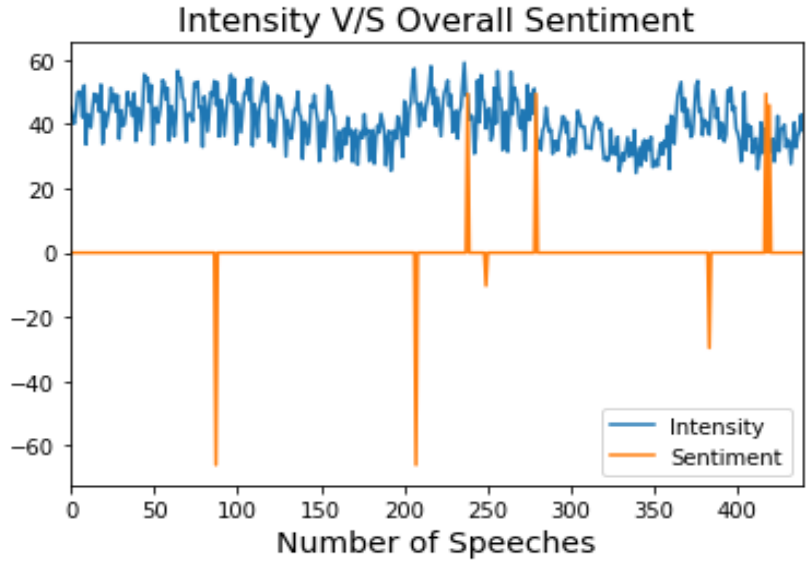

Figure 5.1: Sentiment and Sound Intensity of RAVDESS Audio-Clips

It can be seen that the sentiments and sound intensity are not related in this case. Additionally, most of the audio-clips are neutral in nature. Furthermore, this proves our other finding i.e. sentiment is a better indicator than sound intensity to determine the emotion of speeches.

## 5.4    VidALL NMT Evaluation

This section evaluates the VidALL NMT i.e. the language translation model. In sub-section 5.4.1, the model is evaluated on the basis of their accuracies. In sub-section 5.4.2, the correctness of the translation is evaluated. Finally, in sub-section 5.4.3 a case study is presented where the VidALL NMT is run through the five sentences which were used in Speech Analysis section.

### 5.4.1    Model Evaluation

In Chapter 4, he simple RNN-based and embedded-RNN based models were ran on 10 epochs as an experiment where the latter performed better. For 10 epochs, RNN based and embedded-RNN based models gave accuracies of 61.29% and 82.71% respectively.

To evaluate the model, their performances for 20 epochs were checked. It was found that RNN-based model improved its accuracy to 65.24% whereas the embedded-RNN model achieved an accuracy of 88.71% for 20 epochs.

The validation and training accuracies of the two models were derived for 20 epochs. Figures 5.2 and 5.3 show the training and validation accuracies of the two models which ran for 20 epochs.

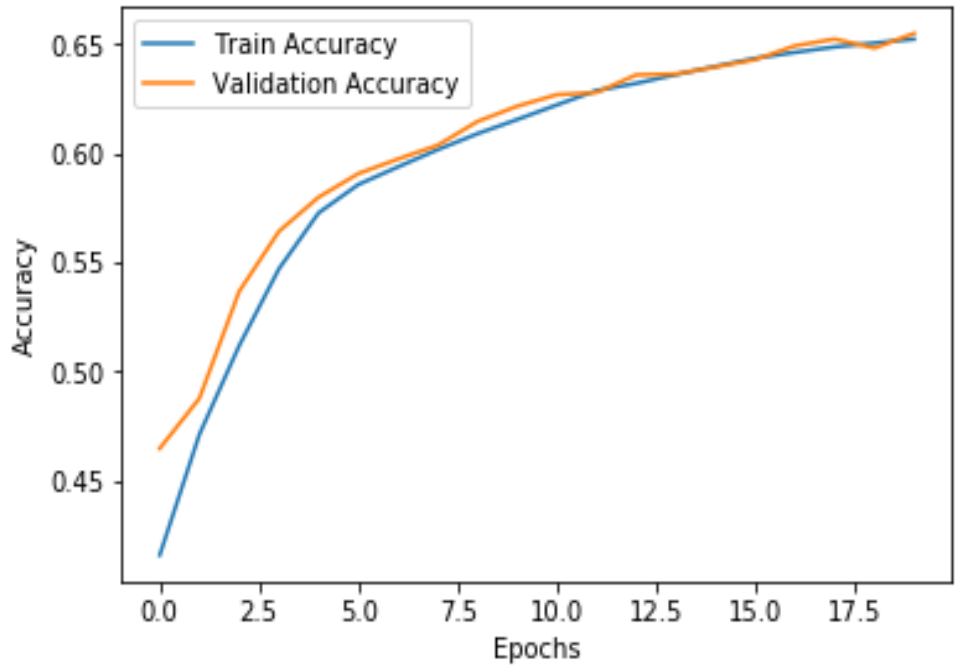

Figure 5.2: Simple RNN-based Model Accuracies for 20 Epochs

From Figure 5.2, it can be seen that the validation accuracy has performed better than the training accuracy for the Simple RNN Model. Despite this, the validation accuracy tried to converge in some points with the train accuracy.

In Figure 5.3, the validation accuracies performed better than the train accuracies throughout the 20 epochs. It tried to converge with the train accuracy towards the end. Thus, it can be stated that both the model performed their best for epochs=20.

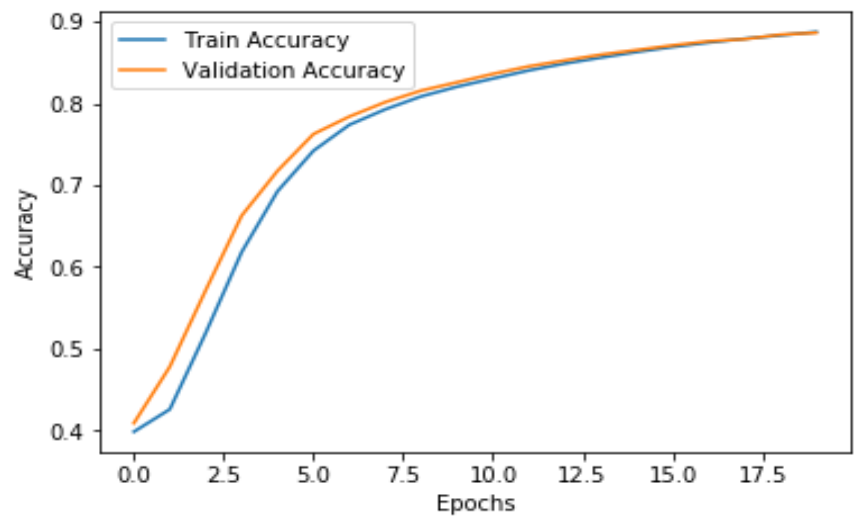

Figure 5.3: Embedded-based RNN Model Accuracies for 20 Epochs

### 5.4.2 Translation Evaluation

The correctness of sentences is a good measure to evaluate the performance of the predicted sentences. The predicted sentences of the two models are compared with the target sentence and translations of Google Translator. The cases for Sentence A and Sentence B are presented in Table 5.1 and Table 5.2 respectively.

| | |
|---|---|
| Source Sentence | *new jersey is sometimes quiet during autumn, and it is snowy in april.* |
| Target Sentence | *new jersey est parfois calmne pendant l' automne, et il est neigeux en avril.* |
| RNN Predictions | *new jersey est parfois parfois en l' et il est est en en.* |
| Embedded-RNN Predictions | *new jersey est parfois calmne en l' et il est neigeux en avril avril.* |
| Google Translator | *le new jersey est parfois calme en automne et il neige en avril.* |

Table 5.1: Sentence A Evaluation of RNN and Embedded-RNN Predictions from 20 Epochs

| | |
|---|---|
| Source Sentence | *he saw a old yellow truck.* |
| Target Sentence | *il a vu un vieux camion jaune.* |
| RNN Predictions | *elle a voiture voiture voiture voiture.* |
| Embedded-RNN Predictions | *il a vu une une camion jaune.* |
| Google Translator | *il a vu un vieux camion jaune.* |

Table 5.2: Sentence B Evaluation of RNN and Embedded-RNN Predictions from 20 Epochs

The BLEU scores were calculated using the NTLK library. For Sentence A, the BLEU scores for the RNN and embedded-RNN predictions were 0.2750 and 0.4020 i.e. they were 27.50% and 40.20% correct respectively. For Sentence B, RNN prediction was not even close to the target sentence, while the embedded-RNN model presented a BLEU score of 0.48 i.e. 48% correctness.

On the whole, it can be inferred that embedded-RNN is a suitable learning model which would provide correct translations.

### 5.4.3 Case Study: Evaluation on VidALL Speeches

| Sentences | Embedded-RNN | Google Translator |
|---|---|---|
| *1. Hello, how are you?* | *l'comment bananes* | *salut comment allez-vous* |
| *2. You know I am finally feeling happy.* | *vous une je je aller aller au* | *vous savez, je me sens enfin heureux* |
| *3. Hi, good to see you.* | *de de de en vous* | *Salut bon de te voir* |
| *4. It feels great to talk to you after such a long time.* | *Il jamais jaimais se en l' vous mois aout d'* | *Cela fait du bien de vous parler après si longtemps* |
| *5. oh my god, look at you.* | *au mon mon est* | *oh mon dieu, regarde toi* |

Table 5.3: Evaluation on VidALL Speech Transcriptions

The embedded-RNN model performed better and gave good translations. Hence, the model was additionally evaluated on the basis of the five sentences from the VidALL Speeches. Table 5.3 presents a comparison of the translated English sentences with the results from the Embedded-RNN model and Google Translator. Evidently, it can be seen that the Embedded-RNN model has failed to provide accurate results.

A good reason behind this could be that the RNN models trained on dataset which contained vocabulary related to third-persons. Evidently, the model has returned results based on its prior learnings. For instance, the model understood that "comment" is a French translation of the word "how". However, it failed to understand the vocabulary in situations where "you" is used.

On the whole, the embedded-RNN model could not yield proper translation predictions for sentences which have been used in VidALL Speeches.

## 5.5  Summary

This chapter discussed about the evaluation strategies for the different phases of this work. In Section 5.2, the VidALL was evaluation on the basis of security and privacy. The motivation behind such a security audit was to ensure user security, data security and determination of any loop holes in the architectural design, Additionally, a human evaluation approach was followed on VidALL as well.

Since the speech analysis of this thesis involved analysis on a smaller number of audio-clips, it was imperative the run the speech analysis model on a larger dataset to validate the findings. Thus, in Section 5.3, a case study was presented where the speech analysis module was evaluated on the RAVDESS audio-clips. The findings were validated after the evaluation in this case.

In Section 5.4, the language translation module was evaluated on basis of model and language correctness. It was found out that embedded-RNNs gave better accuracies, as high as 88.71%, and ensured correctness of predicted of language translations. The language correctness was determined by comparing the predicted sentences with the target sentences and results of Google's language translator. Additionally, BLEU scoring was applied to check the language correctness. To further evaluate the translation model on VidALL speech transcriptions, a case study was presented. A justification was also given on the failure of the translation model on the VidALL transcriptions.

# 6

# Conclusion

Chapter Six ends with a conclusion of this work. In this chapter, the research objectives are assessed individually in Section 6.1. The contribution of this work to the research community has been discussed in Section 6.2. The work faced some difficulties and had limitations which have been discussed in Section 6.3. Finally, in Section 6.4, it has been commented how the work can improve in future.

## 6.1    Assessment of Research Objectives

Chapter Five discusses the various evaluation strategies that have been followed in this work. Since our work contains three phrases, it is important to evaluate each phase based on different parameters. For

### 6.1.1    VidALL:  Audio-Visual Communication System

Research Objective:        *to create an audio-visual communication system which has the ability to record the speech of a speaker*

An important phase of this work was to create an audio-visual application where user can communicate with each other. The whole purpose behind the creation was that it would be easier to run analysis on the speeches of the speakers. Additionally, the need for a real-time language translation in video sessions was realized. Due to the scope and limitations of this work, real-time language translation could not be deployed to VidALL.

The novelty factor of VidALL is that it allows speakers to record their sessions in real-time. Since data security and user-privacy had to be ensured, the speaker can not record the session of the listener i.e. the other user. The audio-visuals were published and subscribed by the users to the OpenTok platform. This was followed by streaming of the audio-visuals between the two user which ensured a proper video communication. With the help of its REST API, the recording of such streams was possible. Although VidALL becomes slow at times, the speakers can hear each other clearly.

### 6.1.2    Speech Corpus Creation

**Research Objective:**        *to create a small speech corpus from the audio-visual communication system*

It was a challenge to find a suitable dataset for analysis of video sessions. The RAVDESS is an audio-visual dataset which contained emotional speeches and videos. However, the audio-visual segments were recorded by individuals. To understand human behaviour in a video chat session, authentic content is needed. Additionally, this objective paved way to solve the need of language translations in video sessions.

The speeches were extracted from the audio-visual files of VidALL where a user can record their messages. Since the recordings were audio-visuals in nature, the audio files were obtained in *.wav* format. The recordings of 20 different individuals were collected in form of audio files and a small corpus was created for our research purpose. Thus, this objective had been met. Additionally, the corpus was used for speech analysis.

### 6.1.3    Behaviour of Speakers

**Research Objective:**        *to understand the behaviour of speakers in a video session*

There was a strong motivation to study and understand the nature of speakers in a video session. This was due to the fact that limited amount of research has been conducted in the field of video chat analysis. A reason for limited research is due to unavailability of video session data as companies ensure user privacy and security. To understand human behaviour in video sessions, a question was formed on how sentiment presence and loudness of voices can result in essential findings.

To meet this objective, voice analysis was conducted on audio-clips of the small corpus which we created. The transcriptions of the audio-clips were obtained via Google's Cloud Speech-to-Text and were used for sentiment analysis. Although loudness of voices was grouped according to the standard comparators, there was no correlation found between acoustic sound intensity and polarity score i.e. the sentiment scores. Additionally, it was concluded that sentiment can be considered as a parameter to understand the human behaviour in a video session. According to some insights, it was observed that individuals tend to start a video session on a positive note. Although, the neutrality of speeches could be due to vocabulary used.

### 6.1.4    VidALL NMT: A Neural Translation Model

**Research Objective:**          *to create a neural translation system to convert English source languages to French*

The recent popularity of neural translation led to several advancements in the field of natural language processing and sentiment analysis. Our work involved creation of a neural based RNN network which trains on data points using GRU. Although the GRUs are being accepted slowly, most RNN models still use LSTM. The challenge was to use GRUs to create an RNN which can train language data on a CPU and give a prediction.

The source sentences passed through the two RNN-based models to predict a target sentence. French and English were chosen as two languages for this purpose as they had similar origin, i.e. Latin. Through extensive experiments it was found out that embedded-RNN based model showed better accuracies than the simple RNN model. It is to be noted that embedded-RNN model used GRU while the simple RNN model used logits in one of the layers. Additionally, the correctness of the predicted sentences was measured using BLEU scores.

### 6.1.5    Speech Corpus in VidALL NMT

**Research Objective:**          *to evaluate and run the translation model with our speech corpus*

The motivating idea behind this objective was a question: Can translated languages occur in real-time video calling? This question gave birth to this objective where a neural model can analyse real-time speeches. For this purpose, the trained neural model was tested on the transcription of the extracted audio-clips. Since the neural model was trained on a dataset which did not contain enough vocabulary, the model could not produce encouraging results from our developed speech corpus.

## 6.2   Contributions

In due course of these extensive work, several findings were gathered which have been explained in various stages. At the same time, there are inferences or observations which can contribute to the research community in some way. The contributions have been briefly discussed below:

- This work produced audio-clip samples from real-time video sessions. This can help a lot of researchers to train their models on real-time audio speeches. However, the number of audio-clips is very small. At the same time, it was due to this corpus, speech analysis on real-life video sessions was possible.

- It was observed that acoustic sound intensity could not be a good estimator in determining the behaviour of individuals in a video session. This will help several researchers as it would help them to save a lot of time.

- This work has resulted in the development of an audio-visual communication system which can record sessions of the speakers. Although the novelty factor of this system can be argued, this can be a good approach for researchers to collect audio clips and work on different research areas.

- This work used GRU instead of LSTM in the RNN layers. Since GRU contain one logic gate less than LSTM in its architecture, the former ensures robustness, scalability and efficiency in a neural network model.

- This work gathered findings from the speech analysis model which suggested that vocal intensities of female speakers tend to be higher than that of the male speakers. Additionally, it was found out that the behaviour of speakers is related to type of vocabulary they use.

On a broad perspective, this piece of work tried to bridge the gap between machine translation and audio-video communication session.

## 6.3   Limitations

The limitations that this work faced or currently possess have been listed below:

- A simple *Speech Recognizer* functionality was used for VidALL to recognize voices of speakers. However, this could not be integrated with VidALL. Since VidALL uses the services of OpenTok cloud platform, the latter prevents any other third party to use the same source of voice. For example, if a user spoke in Bengali, the voice would directly go to the OpenTok platform. Thus, this would prevent the speech to reach to the in-built *Speech Recognizer* function of the android application. Hence, the method had to be scrapped.

- Initially, the recorded sessions are stored online in Amazon S3 bucket for few minutes. An easy method to obtain the videos could have been by web scraping. However, the web address had the presence of algorithmic keys and signatures to ensure user-security. Hence the videos had to be downloaded manually to generate the speech corpus.

- The work could not provide a provide a proper reasoning as to why loudness was picked above other MFCC features for speech analysis. Like in earlier literary works, timbre, pitch and loudness could have been chosen together as parameters for the voice analysis phase.

- The corpora on which the RNN model was trained lacked a strong vocabulary. This was the reason behind the RNN model's inability to produce strong results for the VidALL speech corpus.

## 6.4 Future Work

The prior design approaches and limitations have restricted this project in some ways. At the same time, with a greater number of on-going researches in the field of natural language processing and machine translations, exciting times are ahead. The findings from this work can used to scale up many ideas.

### 6.4.1 Neural Machine Translation for Video Calling

The streaming technology help the developers to make video calling applications. The emergence of research in the video calling applications can open doors to a wide range of opportunities. One such example is to blend neural machine translation and video calling. The learning of live audio-visual sessions will help researchers to make enhanced applications. While real-time language translation is a major research area in academia, live transcriptions in real-time applications have also generated curiosity and paved the way for machine learning approaches in audio-visual applications.

### 6.4.2 Modelling and Training on a GPU environment

This language translation model has been trained on less vocabularies on a TensorFlow-CPU environment. The embedded-RNN based model can be scaled up if the model is trained on a large data set. The non-performance of the model in some cases is due to the lack of training in high vocabulary. Ideally, training RNNs on a larger dataset would require a GPU environment. It would be interesting to see how the different steps would perform if the modelling and training occurred on a GPU environment.

### 6.4.3 Linguists for Evaluation

BLEU scores and Google Translator were used to measure the correctness of the speeches. Since the Google Translator is still re-developing itself, an impartial evaluation approach would be to employ linguists to check the predicted sentences. The human evaluation is one of the standard evaluation processes in machine translation workshops around the world.

Similarly, such measures can improve the evaluation strategy and the correctness of the predicted sentences.